%% file: arxiv_submission.tex
\documentclass{article}

\PassOptionsToPackage{numbers, compress}{natbib}

\usepackage[preprint]{neurips_2020}

\usepackage[utf8]{inputenc} %
\usepackage[T1]{fontenc}    %
\usepackage{hyperref}       %
\usepackage{url}            %
\usepackage{booktabs}       %
\usepackage{amsfonts}       %
\usepackage{nicefrac}       %
\usepackage{microtype}      %
\usepackage{amsmath}
\usepackage{verbatim}
\usepackage{amssymb}
\usepackage{wrapfig}
\usepackage{graphicx}
\usepackage{tabularx} 
\usepackage{multirow}
\usepackage[para]{threeparttable}
\usepackage{subcaption}
\usepackage{bm}
\usepackage{enumitem}

\usepackage{pstricks, pst-node}
\usepackage{verbatim}

\usepackage{tikz}
\usepackage{tikz-cd}
\usepackage{tkz-graph}
\usetikzlibrary{fit,positioning,bayesnet}

\input{defs.tex}

\title{Contrastive Training for Improved\\Out-of-Distribution Detection}

\author{
\parbox{\linewidth}{
\centering
Jim Winkens$^1$, Rudy Bunel$^2$, Abhijit Guha Roy$^1$, Robert Stanforth$^2$, Vivek Natarajan$^1$ \\ Joseph R. Ledsam$^2$, Patricia MacWilliams$^1$, Pushmeet Kohli$^2$, Alan Karthikesalingam$^1$, Simon Kohl$^2$, Taylan Cemgil$^2$, S. M. Ali Eslami$^2$ and Olaf Ronneberger$^2$} \\\\
Google Health$^1$, DeepMind$^2$ \\
\parbox{\linewidth}{
\centering
\texttt{\{jimwinkens, olafr\}@google.com}
}
}

\begin{document}

\maketitle
\begin{abstract}

Reliable detection of out-of-distribution (OOD) inputs is increasingly understood to be a precondition for deployment of machine learning systems. This paper proposes and investigates the use of contrastive training to boost OOD detection performance. Unlike leading methods for OOD detection, our approach does not require access to examples labeled explicitly as OOD, which can be difficult to collect in practice. We show in extensive experiments that contrastive training significantly helps OOD detection performance on a number of common benchmarks. By introducing and employing the \textit{Confusion Log Probability} (CLP) score, which quantifies the difficulty of the OOD detection task by capturing the similarity of inlier and outlier datasets, we show that our method especially improves performance in the  `near OOD' classes -- a particularly challenging setting for previous methods. 

\end{abstract}

\section{Introduction}
\label{sec:intro}
\begin{wrapfigure}[23]{r}{0.5\textwidth}
    \vspace{-1.5cm}
    \centering
    \includegraphics[width=0.46\textwidth]{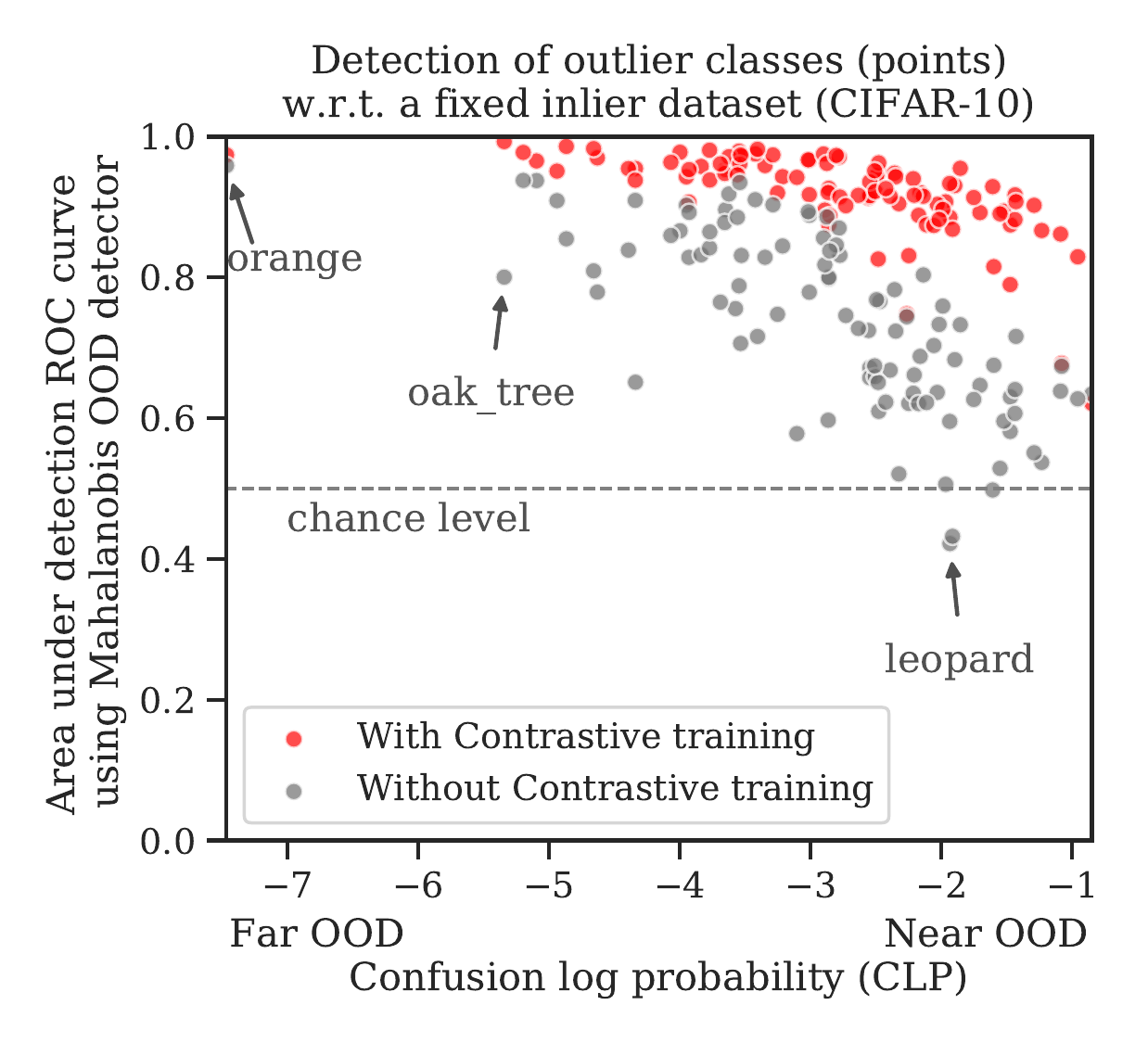}
    \vspace{-0.1cm}
    \caption{
    Each point represents performance at detecting one of the classes in CIFAR-100 as outliers with respect to a network trained on CIFAR-10. The classes are sorted by increasing similarity to the inlier classes as given by the class-wise confusion log probability (CLP), see Eq. \ref{eq:clp}. Contrastive training improves OOD detection results across the board, particularly in the near OOD regime, where the outlier and inlier classes are highly similar. CIFAR-10 contains classes similar to \textit{leopard} (e.g.\ \textit{dog, cat}) but none similar to \textit{oak tree} or  \textit{orange}.}
    \label{fig:fullspectrum}
\end{wrapfigure}

A well-trained deep neural network $f$ that obtains high accuracy on its test set can still make arbitrarily bad predictions when exposed to inputs drawn from an unfamiliar distribution \cite{nguyen2015deep, recht2019imagenet}. This poses a significant obstacle for real world deployment, where it is typically either prohibitively expensive or outright impossible to ensure that the network is only ever exposed to data from the training distribution.

In safety-critical applications, e.g.\ in medical diagnosis, it would be preferable to detect inputs unfamiliar to the trained network for separate processing (for instance by a human expert), than to make potentially inaccurate predictions using the machine learning system. This problem is known as out-of-distribution (OOD) detection, open set recognition, or anomaly detection by different research communities.

Out-of-distribution detection can be performed by approximating a probability density $p(\obs)$ of training inputs $\obs$, and detecting test-time OOD inputs using a threshold $\gamma$: if $p(\obs) < \gamma$ then $\obs$ is considered OOD. The surprising finding of \citet{nalisnick2018deep} was that even powerful neural generative models trained to estimate $p(\obs)$ (e.g.\ on CIFAR-10 images) can perform poorly at OOD detection, often assigning higher probabilities to out-of-distribution test examples (Street View House Numbers) than to in-distribution test examples. 

Modern OOD detection techniques \cite{hendrycks2018deep,hendrycks2019using,lee2018simple,liang2018enhancing} instead assign a scalar score $s(\mathbf{z})$ (e.g.\ via an approximated probability density) to activations $\mathbf{z}$ in an intermediate feature space of a discriminatively trained classifier $f$, and use that to detect OOD inputs. The success of these approaches highly depends on the quality of the intermediate feature space defined by $f$. If the feature space is not sufficiently rich, the network may be blind to properties of the image that turn out to be necessary for detection of OOD inputs. 
Consider, for instance, the case of visual inputs. Variation in captured images is either due to semantic differences of the objects (e.g.\ pose, shape, texture), or due to differences in the imaging process (e.g.\ lighting, camera position). Depending on the application, an unfamiliar variation of either type could lead to an input being deemed out-of-distribution. We therefore desire intermediate feature spaces defined by $f$ that capture as many semantic properties as possible, whilst also remaining sensitive to properties of the imaging process.

Supervised learning produces semantic representations, but only to the extent that those representations discriminate between classes labeled in the dataset. The network $f$ is not incentivized to learn features (semantic or otherwise) beyond the bare minimum necessary to classify. That is why current state-of-the-art approaches to OOD detection \textit{enrich} the intermediate feature space beyond what would ordinarily be learned via only supervised learning on the inlier dataset, for instance by exploiting examples from an outlier dataset (Outlier Exposure, OE, \cite{hendrycks2018deep}), or with self-supervised losses (Rotation Prediction, RP, \cite{hendrycks2019using}). To do this, however, OE requires access to examples labeled explicitly as OOD (which can be difficult to collect in practice), and RP relies on the assumption that an unrelated auxiliary task will produce beneficial representations, which may not always hold.

The key idea of this paper is to encourage $f$ to learn as many high-level, task-agnostic, semantic features as possible from the in-distribution dataset, so as to enable it to detect any kind of out-of-distribution input at test time. We note recently introduced \textit{contrastive} training techniques such as SimCLR \cite{simclr} as performant and well-motivated approaches to this end. Using a set of class-preserving transformations, SimCLR introduces a loss that pulls transformed versions of the same image closer to each other, whilst pushing all other images away. This incentivizes the model to learn features that discriminate between all dataset images, even if they belong to the same class. When combined with supervised training, $f$ learns features that are both rich and semantically discriminative.

\begin{figure}[t]
  \vspace{-0.2cm}
  \begin{minipage}[c]{0.6\textwidth}
    \includegraphics[height=0.53\textwidth]{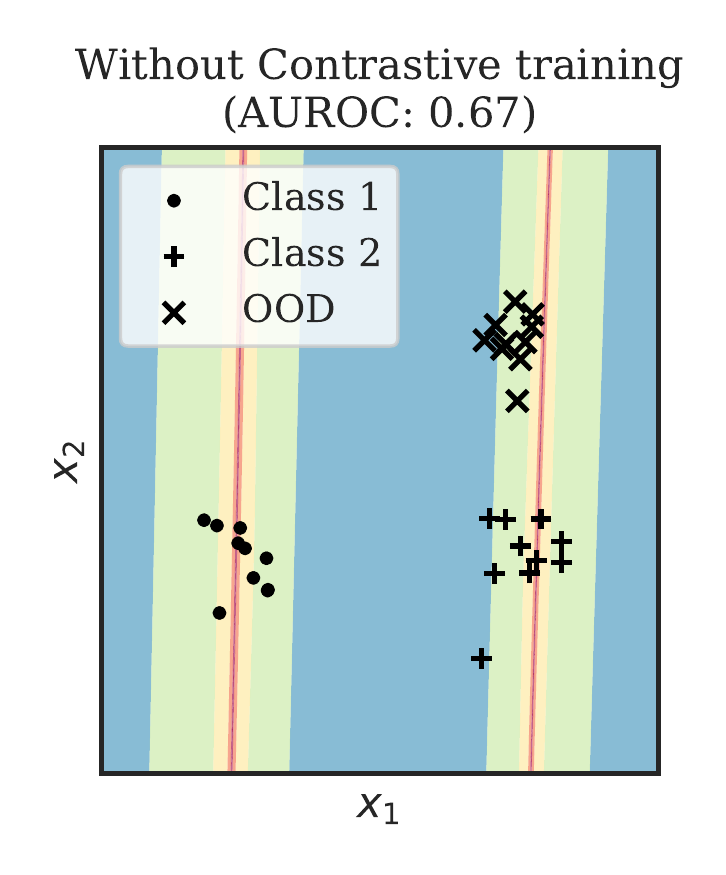}
    \includegraphics[height=0.53\textwidth]{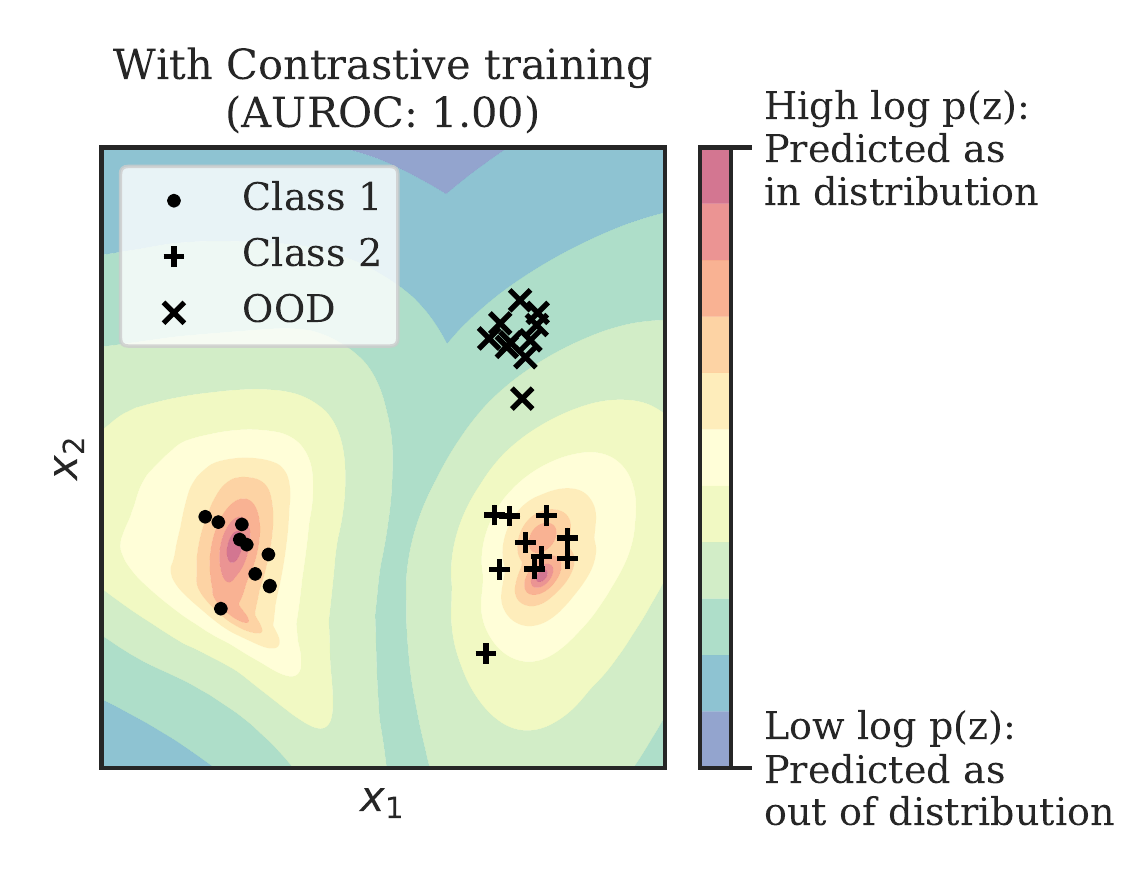}
  \end{minipage}\hfill
  \begin{minipage}[c]{0.3\textwidth}
    \caption{
        We show $s(\mathbf{z}) = \log p(\mathbf{z})$ of a network trained to distinguish between two classes, where $\mathbf{z}$ is its penultimate activation. Left: Supervised training of $\mathbf{z}$ alone may discard input dimensions that are unnecessary for classification but necessary for OOD detection. Right: Contrastive training makes $\lat$ also be sensitive to the $x_2$ dimension.
    } \label{fig:toyexample}
  \end{minipage}
  \vspace{-0.0cm}
\end{figure}

Figure~\ref{fig:toyexample} demonstrates this idea on a toy example, where we aim to classify points in a 2-dimensional input space ($x_1$, $x_2$). The two classes can be distinguished by the first dimension $x_1$ alone. With only supervised training, $f$ has no incentive to be sensitive to the $x_2$ dimension, making OOD detection using $\mathbf{z}$ impossible. Contrastive training, however, shapes $\mathbf{z}$ to remain sensitive to both dimensions and makes OOD detection possible. We show in extensive experiments that this approach scales to high-dimensional problems and consistently improves performance.

An additional difficulty of OOD detection lies in the evaluation of methods. Quantitative evaluation requires specification of an `outlier' dataset at test time, which is itself a subjective design choice, as the notion of `out-of-distribution' is task dependent. We therefore distinguish between \textit{near OOD} regimes where inlier and outlier distributions are meaningfully similar, and \textit{far OOD} regimes where the two are unrelated. Near OOD is encountered more often in practice, e.g.\ a system that detects medical pathologies will often encounter patients with atypical combinations of pathologies (near OOD) and will have to be reliable nonetheless. A completely broken sensor (far OOD) is less prevalent by comparison. We therefore advocate for quantification of the `similarity' of inlier and outlier distributions used in evaluations, and propose a metric for this which we use in our experiments. To summarize, the key contributions of the paper are as follows:
\begin{itemize}[leftmargin=1em,labelwidth=*,align=left]
    \item We propose a new approach for OOD detection that incorporates contrastive training. Our approach avoids explicit inlier and outlier density modelling in the input space, can readily be incorporated into existing training setups and is simple to adapt to different datasets.
    \item We show that the approach consistently improves OOD detection across a wide spectrum of benchmarks, outperforming competitive methods such as Outlier Exposure (OE, \cite{hendrycks2018deep}). Unlike OE, our method does \textit{not} require access to data from the outlier distribution during training or tuning.
    \item We introduce `Confusion Log Probability' (CLP) as a metric to evaluate OOD detection methods, which measures the similarity of inlier / outlier dataset pairs. Using this metric, we show that the proposed method improves OOD detection in both near and far OOD settings, but especially in near OOD settings. See \autoref{fig:fullspectrum} for an overview.
\end{itemize}

\section{Related Work}
\label{sec:related_work}

\textbf{Representations from classification networks.}
The vast majority of OOD detection methods use scores derived from models trained only with multi-class supervision. \citet{hendrycks2016baseline} propose using the maximum softmax probability (MSP) to score OOD samples, which was further improved in ODIN \citep{liang2018enhancing} by using temperature scaling and input pre-processing. \citet{lee2018simple} utilize standard Gaussian density estimates of the network's class-conditional intermediate activations. ~\citet{sastry2019detecting} showed improvements on far OOD detection by using Gram matrices from multiple feature maps.

\textbf{Alternative training strategies.} Beyond proposing better scoring functions, another area of research is adapting the training strategy to improve the quality of scores . MSP scoring was improved by using strategies like confidence loss~\citep{lee2017training}, auxiliary objectives~\citep{devries2018learning, hendrycks2019using, mohseni2020self}, margin loss~\citep{vyas2018out} and outlier exposure~\citep{hendrycks2018deep}. 
A multi-head network architecture was used by~\citet{shalev2018out} to improve the intermediate feature map for OOD detection. Similarly, an approach for novelty detection based on metric learning was presented by~\citet{masana2018metric}. 
Most of the above approaches make the assumption of access to OOD samples during the training process to enhance performance.

{\bf Bayesian approaches.} Under the Bayesian paradigm, \citet{blundell2015weight, malinin2018predictive, chen2018variational} showed that model uncertainty estimates can be produced by learning distributions over network weights and ~\citet{gal2016dropout} proposed Monte-Carlo dropout sampling for it.

\textbf{Generative and hybrid models.} 
An intuitive strategy for detecting OOD samples is to train a generative model from which one can compute the likelihood as an OOD score. An ensemble approach was adapted by~\citet{choi2018waic} and likelihood ratios were estimated as an OOD scoring metric in ~\cite{ren2019likelihood}. While generative models are a promising avenue for OOD detection, applying them directly to the image space has not achieved state of the art results, even when combined with a classification network \cite{nalisnick2019hybrid}. Concurrent to our work, \citet{zhang2020hybrid} show in a surprising  empirical finding that if a residual flow network is attached to the penultimate layer of a classification network, and the networks are trained together, the $p(\mathbf{z})$ learned by the flow network is able to detect near OOD samples much better than baselines in the open set recognition field.

\section{Proposed Method}
\label{sec:replearn}

As shown in Figure~\ref{fig:toyexample}, training using only supervised classification losses may not produce the required features for identifying OOD samples. Contrastive training~\cite{amdim, simclr, moco, cpc} provides a remedy to this problem, by learning a representation capable of distinguishing between all individual training samples, while incorporating existing prior knowledge about identity-preserving transformations.

For image classification, camera parameter and illumination are obvious variations. Both can be approximated by translating, scaling and rotating the image, as well as applying brightness and contrast transformations.
Intuitively, the contrastive loss moves augmented versions of the same image closer together in the feature space whilst pushing all other image pairs apart~\cite{simclr}.

\begin{figure}[htbp]
    \centering
    \includegraphics[width=\linewidth]{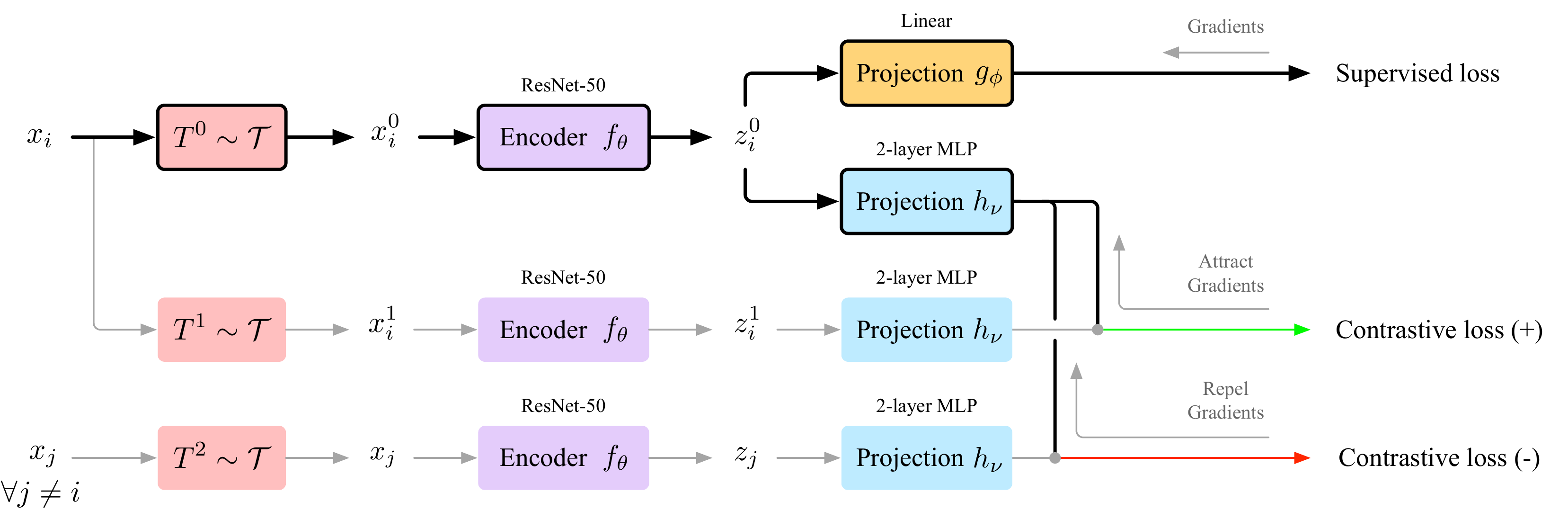}
    \caption{Schematic description of the multitask approach. $\obs_i, \obs_j$: training images. $T$: image transformation (cropping, brightness, etc.). $f_\theta$: encoder network. $\lat$: image represented in latent space. $\projclass$: projection to k classes. $\projcontr$: projection to lower-dimensional embedding space.}
    \label{fig:diagram}
\end{figure}

\textbf{Architecture.}
Our proposed architecture (\autoref{fig:diagram}) for contrastive training is based on SimCLR~\citep{simclr}, chosen for its simplicity.
It is composed of an encoder network $f_\theta$, followed by two projection heads $\projclass$ and $\projcontr$. $\projclass$ maps to the class predictions and $\projcontr$ maps to the low dimensional embedding over which we define the contrastive loss.
The desired feature space is learned on in-distribution training samples only.
For a batch of images \ $\{\obs_i\}_{i=1\dots N}$, we define \mbox{$\obs_i^0 = \trans^0(\obs_i)$} and \mbox{$\obs_i^1 = \trans^1(\obs_i)$}, where $\trans^0$ and $\trans^1$ denote two explicit transformations (such as crop-resize or color distortions) selected randomly from a set of transformations $\mathcal{T}$.
The representations of the transformed images are given as \mbox{$\lat_i^0 = f_\theta(\obs_i^0)$} and \mbox{$\lat_i^1 = f_\theta(\obs_i^1)$} where $f_\theta$ is an encoder, a deep network with parameters $\theta$.

The representation is learned by solving a relational classification task using 
cosine similarity in some embedding space. The cosine 
similarity between any two vectors $\mathbf{u}$, $\mathbf{w}$ is given by \mbox{$\cossim(\mathbf{u}, \mathbf{w}) = \mathbf{u}^\top \mathbf{w} /(\|\mathbf{u}\|\|\mathbf{w}\|)$}.
A 2-layer MLP $\projcontr$ maps a sample from the representation space to a lower-dimensional embedding space $\{\hat{\lat}_i\}$, i.e.\ $\hat{\lat}_i^0 = \projcontr(\lat_i^0)$, and $\hat{\lat}_i^1 = \projcontr(\lat_i^1)$.

\textbf{Objective.}
In this embedding space, the loss function aims to maximize the cosine similarity of sample pairs originating from the same image, i.e.\ $\cossim(\hat{\lat}_i^0, \hat{\lat}_i^1) \rightarrow 1$, whilst minimizing all other pairs originating from two different images, i.e.\ $\cossim(\hat{\lat}_i^a, \hat{\lat}_j^b) \rightarrow 0$, with $j \in \{1, \dots, N\} \setminus i$ and $a,b \in \{0, 1\}$.
The contrastive loss for sample $i$ is then defined as:
\begin{equation}
L_{\text{con}, i} = 
  \sum_{a \in \{0,1\}}
  -\log \frac{
  \exp\left(\cossim(\hat{\lat}_i^a, \hat{\lat}_i^{1-a})/\tau\right)
  }{
  \sum\limits_{j \in \{1, \dots, N\}}
  \!\!\!\!\!\exp\left(\cossim(\hat{\lat}_i^a, \hat{\lat}_j^{1-a})/\tau\right)
  +
  \sum\limits_{j \in \{1, \dots, N\} \setminus i}
  \!\!\!\!\!\exp\left(\cossim(\hat{\lat}_i^a, \hat{\lat}_j^a)/\tau\right)
  }\;,
\end{equation}
where $\tau$ is a positive temperature parameter. The projection head $\projclass$ consists of a linear transformation mapping the representation space to $k$ logits for the $k$ in-distribution classes. The logits are trained with a standard softmax cross-entropy loss $L_\text{class}$.

Training is performed in two stages. The first stage consists of using solely the contrastive loss $L_{\text{con}}$ for a large number of epochs to help learn a good representation. In the second stage, we optimize the combined loss $L_{\text{con}} + \lambda L_{\text{class}}$ for a smaller number of epochs, where $\lambda$ is the supervised loss weight.

\textbf{Density estimation.} We detect OOD samples using a method analogous to \citet{lee2018simple}, by fitting Gaussian distributions to the activations on the training data, which we shape in two significant ways. First, the contrastive loss encourages the network to encode all features capable of distinguishing between samples rather than only those necessary to discriminate between classes. Second, to simplify the distribution of the activation that is fitted, label smoothing is added to the cross-entropy loss $L_{\text{class}}$, so as to prevent the network from spreading out the activations in an attempt to drive the logits of the correct class to infinity. This encourages tight clustering of the activations of each class, as demonstrated by \citet{muller2019does}. 

To take advantage of this last property, our density estimation is performed class-wise, over the activations $\lat$ at the penultimate layer. For each class $c$, we estimate an $n$-dimensional multivariate Gaussian $\mathcal{N}(\bm{\mu}_c,\mathbf{\Sigma}_c)$, with $n$ the dimension of $\mathbf{z}$. For the OOD score $s(\obs)$, the highest density is taken over all the class-conditional Gaussian components. A high score $s(\obs)$ indicates that the representation of a test sample in the embedding space lies close to the typical set for one of the classes. Conversely, a low score signifies that the test sample has a representation that is far from all training set examples and is therefore likely to represent an OOD example.
\begin{align}
    & s(\obs) = \max_{c} \left[ - (f_\theta(\obs) - \bm{\mu}_c)^T \mathbf{\Sigma}_c^{-1} (f_\theta(\obs) - \bm{\mu}_c) - \log\left((2\pi)^{n} \det \mathbf{\Sigma}_c \right) \right],
    \label{eq:scoring}
\end{align}
where $\bm{\mu}_c$ and $\mathbf{\Sigma}_c$ are obtained using the standard estimators.
Unlike \citet{lee2018simple}, we do not perform ensembling over detectors based on different layers, since determining the optimal linear combination would require access to labelled OOD data. We also do not perform input preprocessing~\citep{liang2018enhancing}. Instead, we rely on having a richer representation over which to define our distribution.

\section{Confusion Log Probability (CLP) as a Measure of Dataset Distance}
\label{sec:similarity}

Real world test samples may vary strongly, from falling exactly within the training distribution to falling far out of the training distribution. As a consequence, a robust model should exhibit strong OOD detection performance across the entirety of the spectrum. Current benchmarks however report only a single performance statistic for each in-distribution and OOD dataset pair, such as the area under the receiver operating characteristic curve. This type of evaluation metric can therefore not resolve the performance in detecting near versus far OOD.

Different proposals toward discerning the difficulty of OOD tasks have been made in the literature. The openness score~\cite{scheirer2012toward}, used in the Open Set Recognition community, gives a difficulty measure based on the number of classes in the training set compared to the number of classes in the test set. This measure, however, ignores the visual similarity that can exist between classes, and therefore the relative difficulty of detecting different unknown classes. Another example is the usage of the maximum mean discrepancy with an $L^2$ distance kernel on the image space~\cite{liang2018enhancing}. However, applying the $L^2$ distance directly in the image space can only identify nearly identical images. Therefore this metric cannot completely assess the spectrum of OOD cases which we seek.

We propose the \textbf{confusion log probability (CLP)} as a measure that is indicative of the difficulty of an OOD detection task. CLP is based on the probability with which a classifier confuses outliers with inliers, that has access to outlier samples during training. Given two labelled datasets $\Data_{\text{in}}$ and $\Data_{\text{out}}$, with corresponding sets of classes $\mathcal{C}_{\text{in}}$ and $\mathcal{C}_{\text{out}}$, we train an ensemble of $N_e$ classifiers $\{\phat^j\}_{j=1}^{N_e}$ on the joint dataset $\Data = \Data_{\text{in}} \cup \Data_{\text{out}}$ using the extended label set  $\mathcal{C} = \mathcal{C}_{\text{in}} \cup \mathcal{C}_{\text{out}}$. Once the ensemble is trained, we compute an estimate for the confusion matrix between classes on held-out test data. The expected probability of a test sample $\obs$ to be predicted as class $k$ is given by:
\begin{equation}
c_k(\obs) = 
\frac{1}{N_e}
\sum_{j=1}^{N_e} \phat^j(\hat{\cls} = k | \obs).
\label{eqn:simm_score}
\end{equation}
Therefore the confusion of a set of test OOD samples $\mathcal{D}_{\text{test}}$ with the inlier classes $\mathcal{C}_{\text{in}}$, i.e.\ the confusion log probability (CLP) of $\mathcal{D}_{\text{test}}$, becomes:
\begin{equation}\label{eq:clp}
        \mathrm{CLP}_{\mathcal{C}_{\text{in}}}(\mathcal{D}_{\text{test}}) = 
        \log \left(
    \frac{1}{|\mathcal{D}_{\text{test}}|}
    \sum_{\obs \in \mathcal{D}_{\text{test}}} 
    \sum_{k \in  \mathcal{C}_{\text{in}}} 
    c_k(\obs)
    \right).
\end{equation}

A low CLP indicates that test samples are far OOD and a high CLP indicates that they are near OOD. Note that summing over the probability for all inlier classes $\mathcal{C}_{\text{in}}$ effectively evaluates the binary classification problem of inlier versus outlier classes on outlier samples. As such CLP is asymmetric.

We can compute a class-wise CLP with $\mathcal{D}_{\text{test}}$ being the test samples of this specific class, or a dataset CLP by using all samples of the test dataset as $\mathcal{D}_{\text{test}}$. The class-wise CLP is used in Fig.~\ref{fig:fullspectrum}, where we show how OOD detection performance varies as a function of the distance between the CIFAR-100 classes and the inlier CIFAR-10 dataset. 
By using classifiers to estimate class confusions ($5$ ResNet-34 models in our work), we ground the measure in a notion of visual similarity rather than semantic or image space similarity. 
The choice of an ensemble of independently trained classifiers is motivated by their well-calibrated predictions over single classifiers \cite{lakshminarayanan2017simple}. Further implementation details of the ensemble training setup to compute CLP and further qualitative analysis using dendrograms can be found in Appendix~\ref{appx:dendogram}.

\section{Experiments}
\label{sec:experiments}

\textbf{Near to far OOD spectrum benchmark.} We study out-of-distribution detection on the following in-distribution dataset ($\Data_{\text{in}}$) and out-of-distribution dataset ($\Data_{\text{out}}$) pairs which we believe represent the most challenging pairs of common OOD detection benchmarks \citep{hendrycks2016baseline}: We use the CIFAR-10 and CIFAR-100 \citep{Krizhevsky09learningmultiple} datasets, as well as the Street View House Numbers (SVHN) dataset \citep{Netzer2011}. Note that the CIFAR-10 and CIFAR-100 classes are mutually exclusive. The distance of the dataset pairs is given by the min-max bounds of the class-wise CLP.

\textbf{Evaluation metrics.} As is common in the OOD detection literature, we use the area under the receiver operating characteristic curve (AUROC), with in-distribution and out-of-distribution being the labels, as the metric for OOD detection performance. Note that this is a threshold-independent and calibration-free evaluation metric. Results with additional metrics can be found in Appendix~\ref{appx:additional_res}.
In addition to AUROC, we also quantify the performance on a single OOD sample using the \textit{OOD rank}. For estimating this, we compare the score of the OOD sample $s(\obs)$ to the scores of the inlier test dataset. The OOD rank is given by the percentage of inlier test examples that have a lower $s(\obs)$ than the OOD sample. The higher the rank, the more the sample is deemed to be OOD.

\begin{table}[t]
\centering
\begin{threeparttable}

\caption{Out-of-distribution detection performance (AUROC).} 
\label{tab:mainresults}

\begin{tabular}{lllll}\toprule
\multirow{ 4}{*}{\textbf{Method}}             & \textbf{Near OOD} & \textbf{Near \& Far OOD} & \textbf{Far OOD} & \multirow{ 4}{*}{\textbf{Average}}   \\ 
 &$\Data_{\text{in}}$ = CIFAR-100      &  $\Data_{\text{in}}$ = CIFAR-10           & $\Data_{\text{in}}$ =   CIFAR-10\\
 & $\Data_{\text{out}}$ = CIFAR-10      &  $\Data_{\text{out}}$ = CIFAR-100        &  $\Data_{\text{out}}$ = SVHN &    \\
 &\small CLP= [-$4.5$ to -$2.6$] &\small CLP=[-$7.4$ to -$0.8$] &\small CLP=[-$12.1$ to -$7.6$] &\\
 
              \midrule

Softmax probs. \citep{hendrycks2016baseline}                 &   77.1                 & 86.4                  & 89.9   &      84.5      \\
ODIN \citep{liang2018enhancing}*         &     77.2               &   85.8                 &    96.7           & 86.6 \\
Mahalanobis \citep{lee2018simple}*        &     77.5             &  88.2                  &   99.1            & 88.3\\
Residual flows \citep{zisselman2020deep}*      &   77.1                 & 89.4                   &     99.1         &  88.5\\
 
Outlier exposure  \citep{hendrycks2018deep}$^{\dagger}$ &  75.7$^{\dagger}$                  &      \textbf{93.3}$^{\dagger}$              &  98.4$^{\dagger}$      &      89.1$^{\dagger}$  \\

Rotation pred.  \citep{hendrycks2019using}$^{\ddag}$ &  -                  &      90.9              &  98.9        &      -  \\

Gram matrix \citep{sastry2019detecting}         &   67.9                 &                  79.0  &     \textbf{99.5}         & 82.1 \\
 \textbf{Ours} & \textbf{78.3}       &                  92.9  &  \textbf{99.5}    &      \textbf{90.2 } \\
 
\bottomrule

\end{tabular}

\begin{tablenotes}\small
\item[*] \small{Uses data explicitly labeled as out-of-distribution for tuning}
\item[\textdagger] \small{Uses data explicitly labeled as out-of-distribution for training}
\item[\textdaggerdbl] \small{Uses additional data for pretraining}
\end{tablenotes}
\end{threeparttable}

\end{table}

\textbf{Setup.} We adopt a wide ResNet-50 \citep{he2016deep} as the encoder $f_\theta$, whose last layer has a fixed-dimensional output (6144-D). This output vector $\mathbf{z}$ is the representation  on which we compute the  OOD score $s(\obs)$ for a given test sample $\obs$.  $\mathbf{z}$ is followed by
the supervised head $\projclass$ with label smoothing, and the contrastive head $\projcontr$ with a 128-D 2-layer MLP projection with batch normalization and ReLU.  For fair comparison, we use the same architecture, the same transformation function $\mathcal{T}$ and scoring function for the baselines without contrastive training reported in \autoref{tab:trainingmodes}. Further training setup details are provided  in Appendix~\ref{appx:implementation}. 

\begin{figure}[b]
    \begin{subfigure}{0.33\textwidth}
    \vspace{-5pt}
    \includegraphics[width=\linewidth]{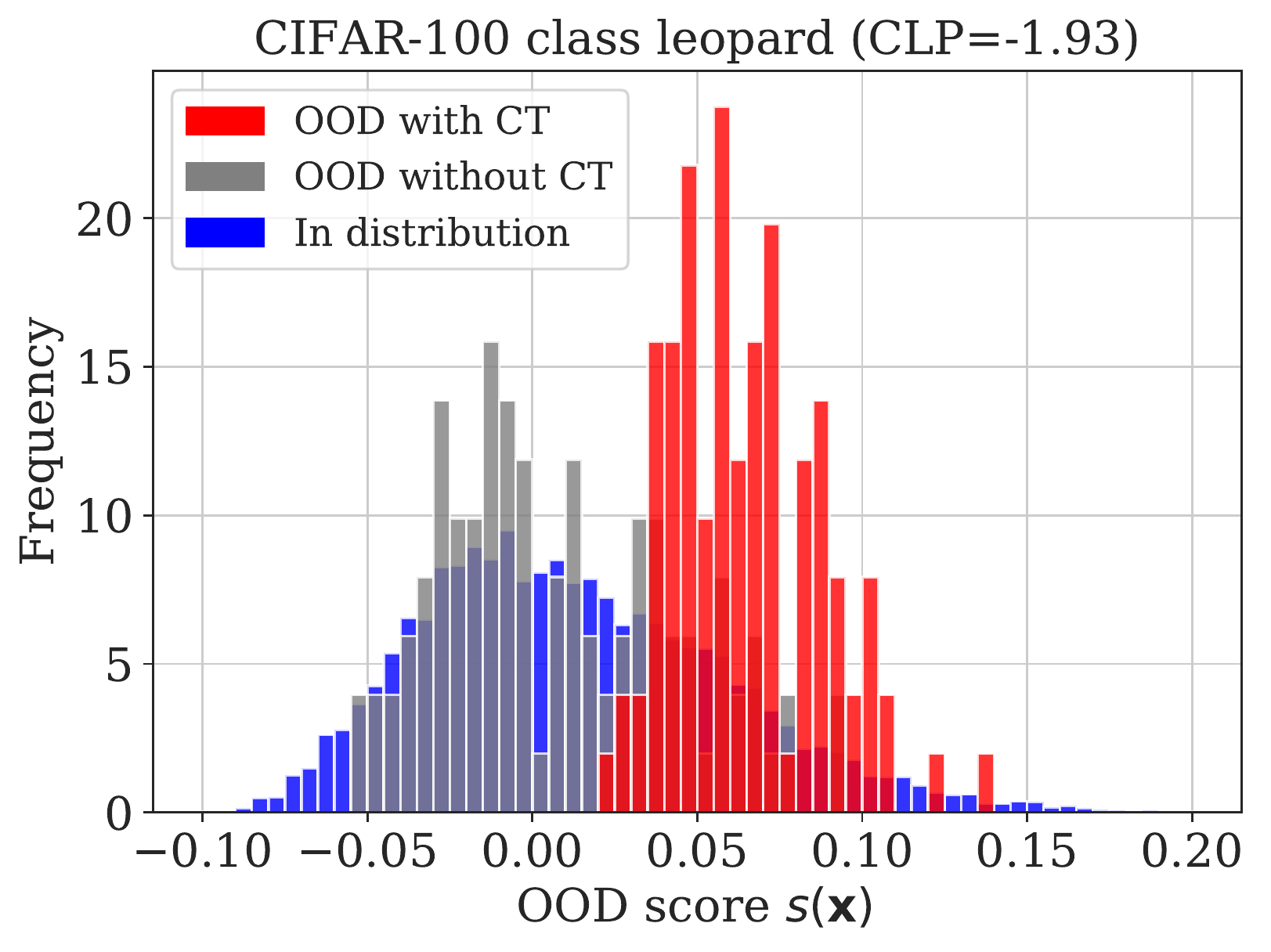}
    \caption{Near OOD}
  \end{subfigure}%
  \hspace*{\fill}   %
  \begin{subfigure}{0.33\textwidth}
  \vspace{-6pt}
    \includegraphics[width=\linewidth]{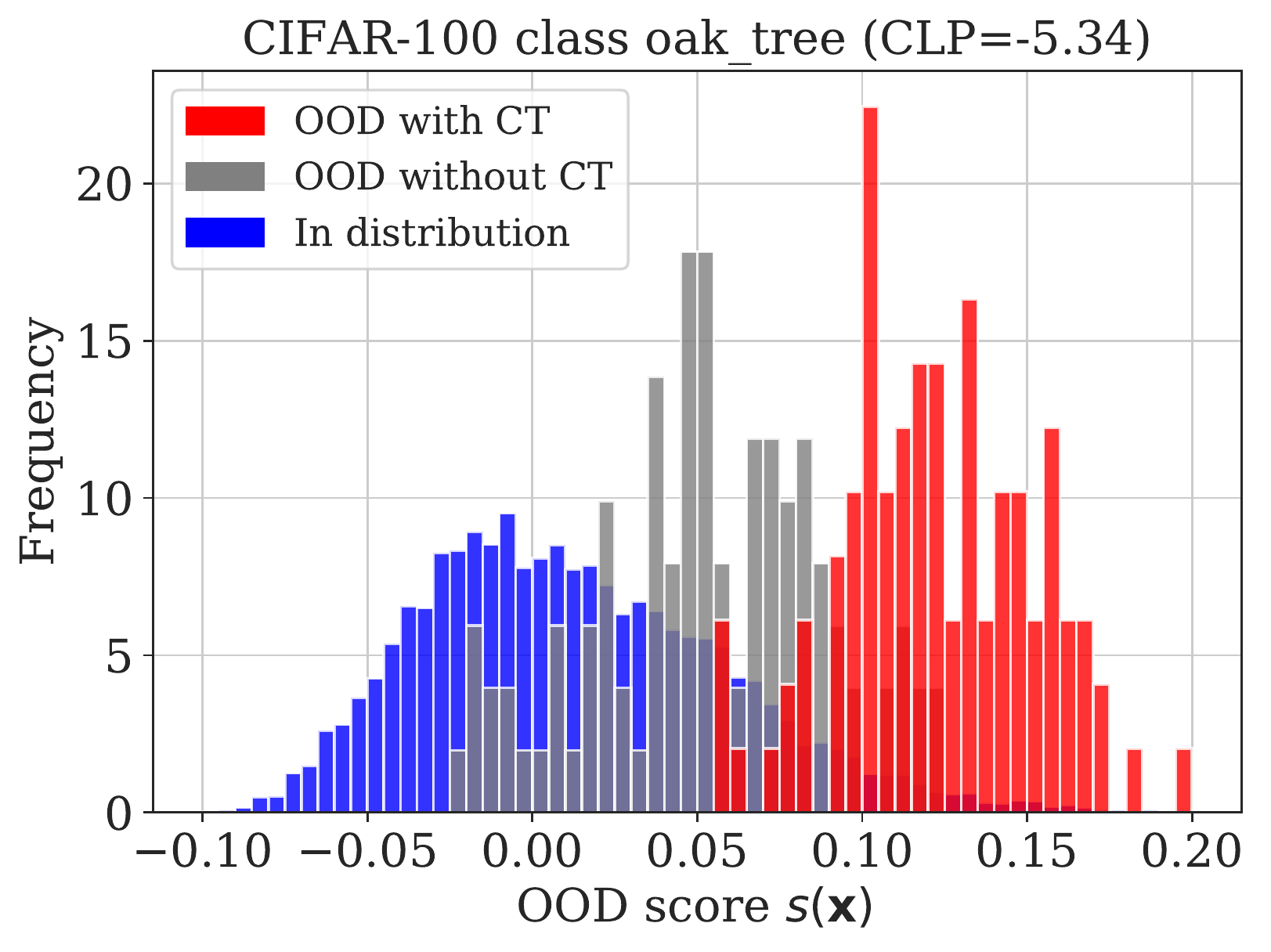}
    \caption{Far OOD}
  \end{subfigure}%
  \hspace*{\fill}   %
  \begin{subfigure}{0.33\textwidth}
    \includegraphics[width=\linewidth]{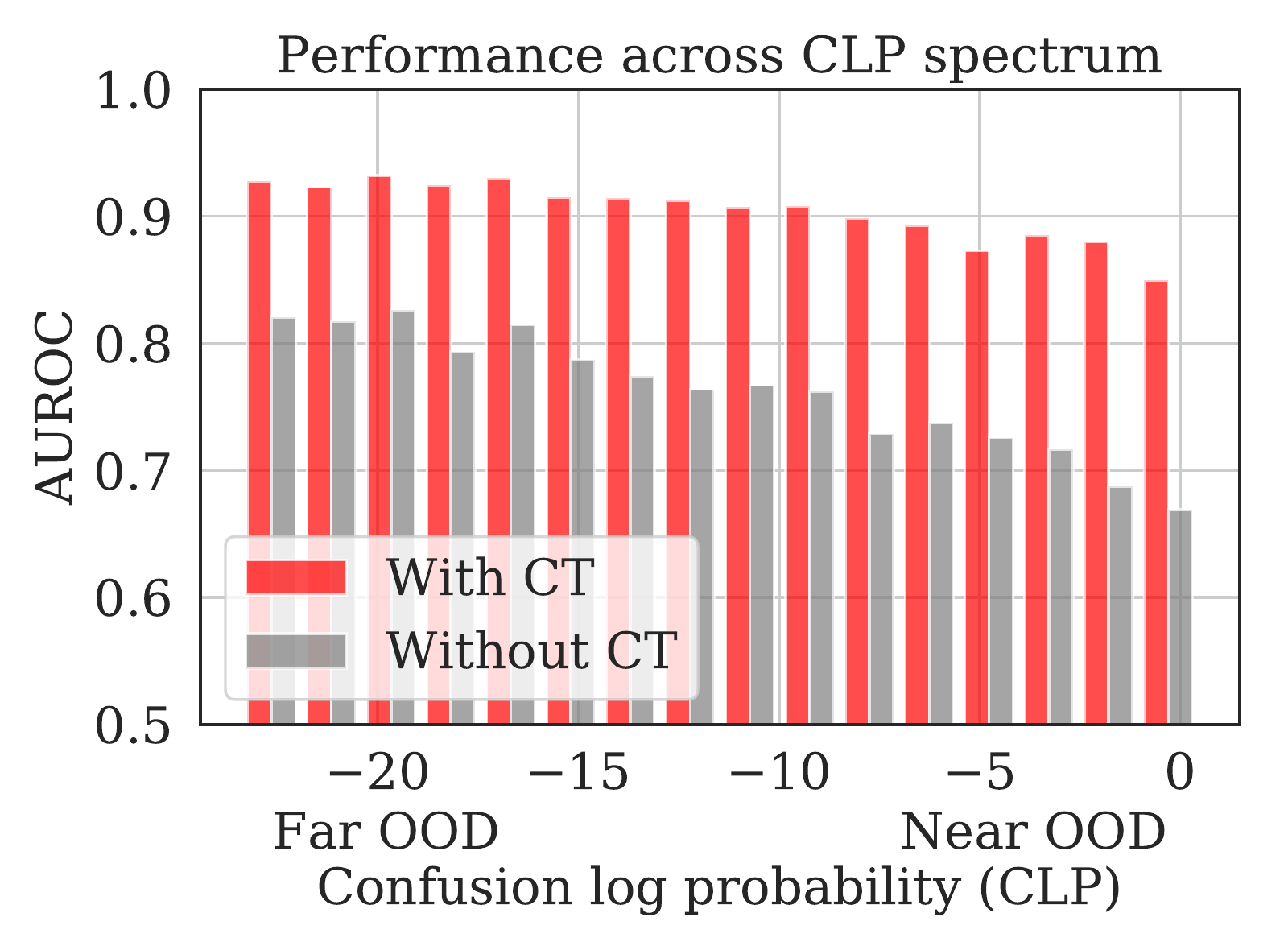}
    \caption{CLP Spectrum}
  \end{subfigure}
  \hfill
  \caption{Experimental results with contrastive training (CT) under the $\Data_{\text{in}}$ = CIFAR-10, $\Data_{\text{out}}$ = CIFAR-100 setting, showing the histograms of OOD scores $s(\obs)$ with respect to those of a fixed inlier dataset for \textbf{(a)} a near OOD class and \textbf{(b)} a far OOD class.  \textbf{(c)}: Out-of-distribution detection performance across the sample-wise CLP spectrum.}
  \label{fig:histograms}
\end{figure}

\textbf{Results.} \autoref{tab:mainresults} shows our main results. We observe that our proposed method improves OOD detection across the spectrum of our proposed benchmark. Specifically, on the near OOD dataset pair with CIFAR-100 as $\Data_{\text{in}}$ and CIFAR-10 as $\Data_{\text{out}}$, we obtain an AUROC of 78.3 which is, to the best of our knowledge, a new state-of-the-art result, outperforming \citep{lee2018simple}. With CIFAR-10 as $\Data_{\text{in}}$ and CIFAR-100 as $\Data_{\text{out}}$, containing both test samples that are near as well as far OOD, we obtain an AUROC of 92.9, which is more than two points better than the next best method not using labeled out-of-distribution data during training. \autoref{fig:histograms}a shows contrastive training helps to differentiate OOD classes that are highly similar to inlier classes, where the baseline results is worse than random performance. In the far OOD setting (\autoref{fig:histograms}b), contrastive training is also a significant component for further separation between $\textbf{z}$ of dissimilar images and inlier dataset images. The performance gap is the largest for the high CLP regime (+18 AUROC points for $-5\leq \text{CLP} <0$), and remains for the whole spectrum (\autoref{fig:histograms}c). On the far OOD dataset pair with CIFAR-10 as $\Data_{\text{in}}$ and SVHN as $\Data_{\text{out}}$, we obtain an AUROC of 99.5 which is on par with the current state-of-the-art \citep{sastry2019detecting}.  Results on additional dataset pairs are reported in Appendix~\ref{appx:additional_res}.

When considering the average performance on all three dataset pairs as representative of the entire near to far OOD spectrum, our method obtains an AUROC of 90.2 outperforming the previous state-of-the-art method \citep{hendrycks2018deep}. It is worth pointing out that, unlike others, our approach \textit{does not} assume access to additional data or labels from an outlier distribution during training or tuning.

\begin{table}[t]
  \caption{\textbf{Ablation study  of objective function.} The baseline model only performs supervised training and we evaluate the impact of incorporating label smoothing (LS) and contrastive training (CT). We report AUROC, as well as the standard deviation of the OOD rank between 5 runs.}
\label{tab:trainingmodes}
\small
\centering
\begin{tabular}{c@{\hskip -0.3in}c@{\hskip -0.3in}ccc@{\hskip 0.3in}ccc}
\toprule
  \multicolumn{2}{c}{\multirow{2}{*}{\begin{tabular}{c}\textbf{Training}\\\textbf{strategy}\end{tabular}}} & \multicolumn{3}{c}{\textbf{AUROC}} &\multicolumn{3}{c}{\multirow{2}{*}{\begin{tabular}{c}\textbf{OOD rank}\\\textbf{Average standard deviation}\end{tabular}}}\\
  \\
  \cmidrule(r{1em}){3-5} \cmidrule{6-8}
  \multirow{2}{*}{LS} & \multirow{2}{*}{CT} & $\Data_{\text{in}}$ = CIFAR-100 & CIFAR-10 & CIFAR-10 & CIFAR-100 & CIFAR-10 & CIFAR-10\\
  & & $\Data_{\text{out}}$ = CIFAR-10 & CIFAR-100 & SVHN & CIFAR-10 & CIFAR-100 & SVHN\\
  \midrule
  x & x & $63.9 \pm 0.3$  & $81.1 \pm 0.2$ & $96.7 \pm 0.2$ & 23.3 & 20.7 & 6.7\\
  \checkmark & x & $74.1 \pm 0.4$  & $90.8 \pm 0.1$ & $99.2 \pm 0.1$ & 20.3 & 12.8 & 2.0\\
  x & \checkmark & $72.1 \pm 0.4$ & $90.9 \pm 0.2$  & $98.8 \pm 0.2$ & 23.7 & 12.5 & 2.4\\
  \checkmark & \checkmark & $78.3 \pm 0.2$ & $92.9 \pm 0.2$ & $99.5 \pm 0.1$ & 19.3 & 10.6 & 1.7\\
  \bottomrule
\end{tabular}
\end{table}

\textbf{Impact of activation space shaping.} 
We perform an ablation study to investigate the impact of label smoothing and contrastive training, with results reported in \autoref{tab:trainingmodes}.
We report results at the task level, but also investigate the variation on a per-sample basis, by looking at the OOD rank variation between several randomly initialized runs.
This allows us to quantify whether the OOD samples considered in distribution always remain the same or whether the errors vary. 

We observe that while using contrastive training or label smoothing alone leads to improvements over the supervised baseline, significantly better results are obtained when the two of them are combined together to shape the activation space. We hypothesize that scoring with standard Gaussian density estimation benefits significantly from tighter class clusters obtained via label smoothing, and it is required for the combination of contrastive training and the scoring function as seen in Eq. \ref{eq:scoring} to work effectively. Furthermore, our feature shaping strategies allow for a significant reduction in variation of the OOD rank of samples between runs. We hypothesize that for the supervised baseline, only features required for class discrimination will be created while the presence of task-agnostic features useful for near OOD detection will only be present by chance due to randomness in training.

More details on additional ablation tests and discussions are available in Appendix~\ref{appx:additional_ablation}, where we study the impact of parameters like relative weighting between supervised and contrastive loss, temperature and model capacity on OOD detection performance.

\begin{figure}[t]
    \begin{subfigure}{0.43\textwidth}
    \includegraphics[height=0.75\linewidth]{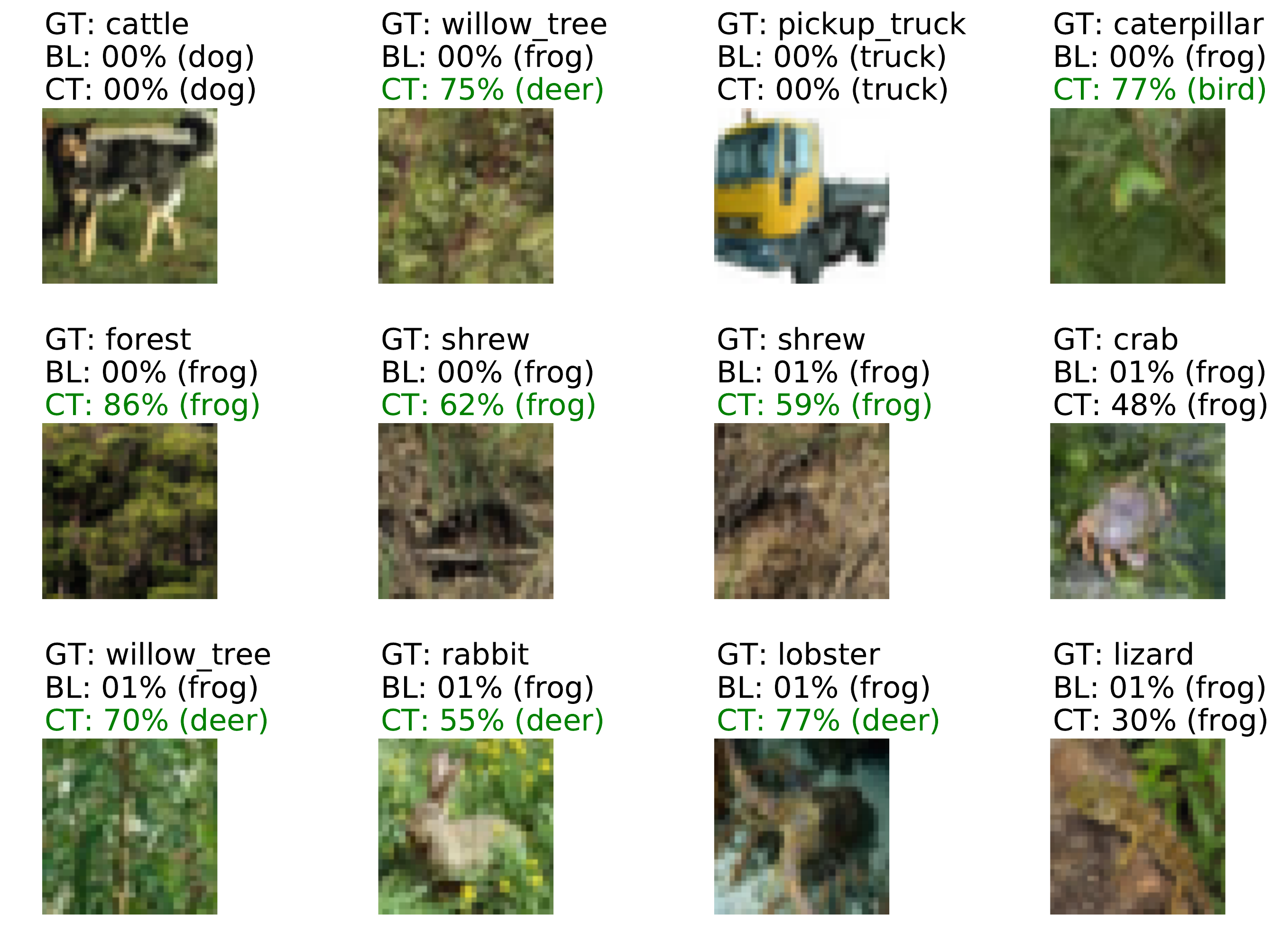}
    \caption{Worst mistakes by baseline (BL)} \label{fig:1a}
  \end{subfigure}%
  \hfill
  \begin{subfigure}{0.43\textwidth}
    \hfill
    \includegraphics[height=0.75\linewidth]{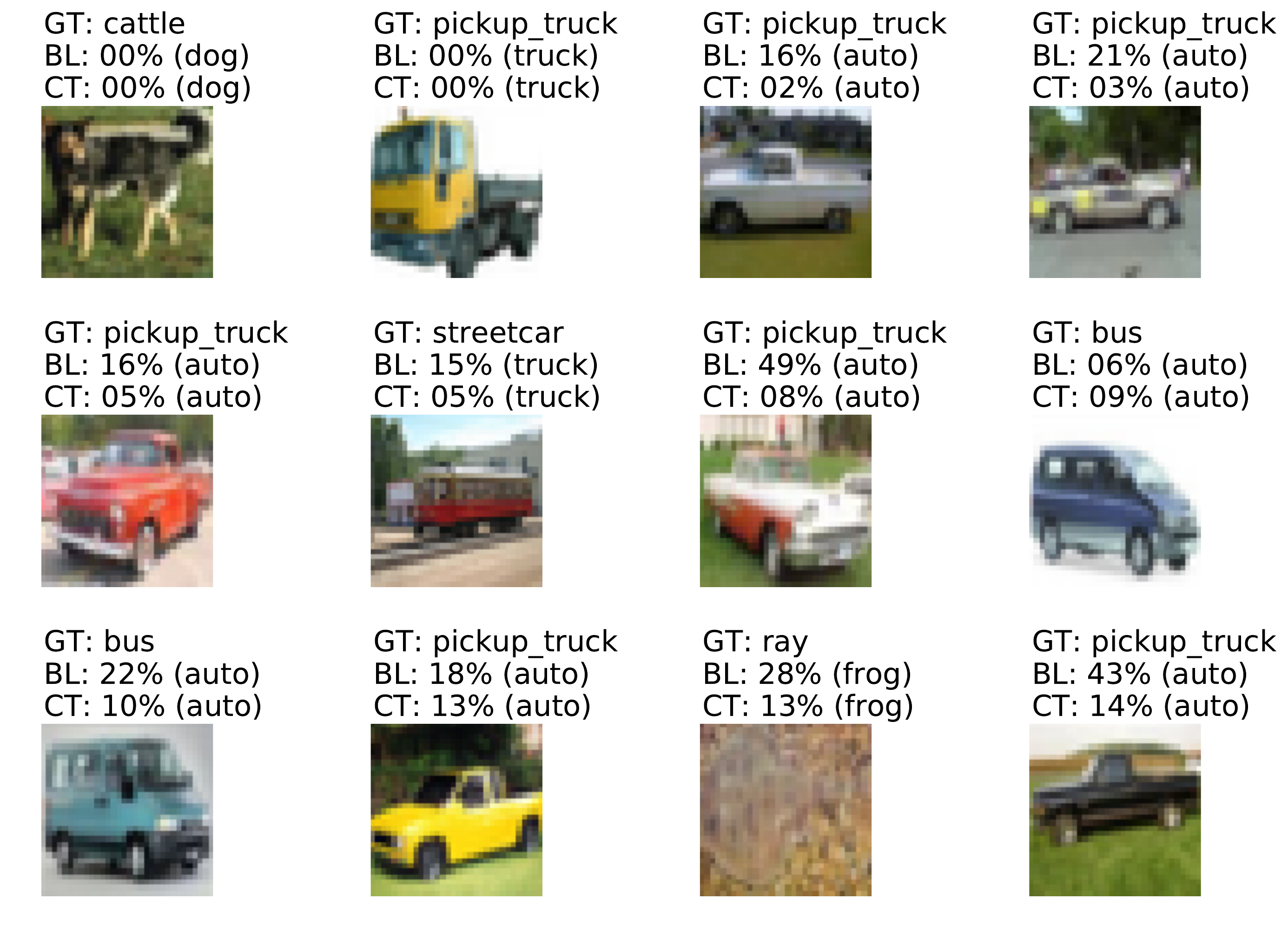}
    \caption{Worst mistakes by contrastive training (CT)} \label{fig:1b}
  \end{subfigure}
  \caption{\textbf{Failure mode analysis.} $\Data_{\text{in}}$ = CIFAR-10; $\Data_{\text{out}}$ = CIFAR-100; \textbf{GT}: CIFAR-100 ground truth label. \textbf{BL}: baseline, \textbf{CT}: contrastive training. The numbers  show the percentile rank of the OOD score. In brackets the class of the most activated inlier Gaussian. Green: percentile rank > 50\%, indicating relative success of a method at detecting that input as an outlier.}
  \label{fig:failuremodeanalysis}
\end{figure}

\textbf{Failure mode analysis.} \autoref{fig:failuremodeanalysis} shows failure cases for the baseline and our model. We show samples from $\Data_{\text{out}}$ = CIFAR-100 most likely to be \textit{incorrectly} predicted as in-distribution for networks trained on $\Data_{\text{in}}$ = CIFAR-10. The percentile rank of the OOD score w.r.t. the scores of the inlier dataset indicates the success of the OOD detection (higher is better). For most of the baseline's worst mistakes, our approach succeeds in identifying that sample as an outlier.

\section{Discussion and Conclusion}

In this paper, we proposed a simple contrastive training-based approach to OOD detection that outperforms recent methods across a variety of settings. 
Furthermore we introduced a new metric to quantify the difficulty of a specific OOD task, the confusion log probability (CLP), that captures the similarity of inlier and outlier datasets. We showed with extensive experiments that representations obtained through contrastive training improve OOD detection performance beyond what is possible with purely supervised training. The representations are shaped by joint training, in which the contrastive loss pushes the representations apart, even within each class, while the supervised loss acts to cluster the representations by class. Unlike previous competitive methods our approach does not require access to data from the outlier distribution during training or tuning. Moreover, in contrast to many other representation learning approaches, the underlying SimCLR \cite{simclr} has a well-motivated training objective, and can scale to large images and datasets.

Of course this additional training objective can not guarantee that all necessary features for near OOD classes are found. See \autoref{fig:failuremodeanalysis} for a failure mode analysis for the setting $\Data_{\text{in}}$ = CIFAR-10 and  $\Data_{\text{out}}$ = CIFAR-100. Here the model saw `automobiles' and `trucks' during training, but it was not able to distinguish the new classes `pickup truck' and `bus' from the existing ones. These class-pairs are however also the most challenging for fully supervised training as CLP shows.

This work only examines improvements that arise from training a richer representation $\lat$. With regards to scoring of $\lat$, we employed standard Gaussian density modeling like prior methods \cite{kamoi2020mahalanobis}. In concurrent work, \citet{zhang2020hybrid} show that scoring with a deep flow-based network can also significantly improve performance. We expect that these improvements in density estimation are complementary to our proposed contrastive training. 

We use a larger neural network than is typical in the literature. This capacity is needed to encode the additional features that SimCLR training creates. We have shown that the Mahalanobis approach \cite{kamoi2020mahalanobis} does not benefit from additional model capacity.
For other baselines we can only speculate that the authors tried to add more capacity, and published the network that yielded best performance.

Another advantage of our setup is the ability to extract useful information from completely unlabelled images that can come from arbitrary distributions. Other approaches that learn a density directly from the training set (e.g.\ \citet{zhang2020hybrid}) rely on the fact that all training examples come only from the `in-distribution' set.
Especially in real-world applications, for instance medical imaging, a large set of unlabelled images is easily available from routine imaging. Removing outlier images from this set would be an expensive manual labelling task.

All in all, this work demonstrates that the challenges in OOD detection (especially for the near OOD setting) are closely related to those in unsupervised representation learning. We believe that viewing the OOD detection problem from this perspective opens up new avenues for progress.

\section*{Broader Impact}

Deep neural networks have demonstrated superhuman performance across a wide range of applications, with the promise of significant positive impact on the world. Despite this, our ability to safely deploy models in real world settings is still limited. One such obstacle is the inability of existing models to accurately withhold prediction for an input that is meaningfully different from typical examples encountered during training. In contrast to human experts, deep neural network based systems tend to struggle to account for the uncertainties for taking an appropriate decision, e.g.\ refer a difficult case for a second opinion. Moreover, models can fail in unexpected ways, making errors difficult to identify in practice. In domains with the greatest potential for societal impact, such as medical imaging and self-driving cars, appropriate recognition of OOD inputs is essential to avoid catastrophic errors that may cause harm.

This work proposes a new scalable approach to OOD detection based on recent advancements in representation learning, and takes a step towards the ultimate goal of enabling safe real-world deployment of machine learning models in safety-critical domains. Additionally, this work also defines and recognizes the importance of evaluation of OOD detection performance in a spectrum  of regimes, in particular near and far OOD settings. The experimental results have been reported on standard benchmark 
datasets for considerations of reproducible research, but both the OOD detection method and the evaluation approach are addressing potential issues encountered in safety critical domains. 
For example, in medical imaging, pathologies, poor-quality images and other artefacts that have not previously been encountered in model training are both common and important. 
Progress in methods for near OOD detection is therefore particularly relevant to addressing the often subtle but significant challenges of operating safely in a real world environment.

\bibliography{references}\bibliographystyle{icml2019}
\vfill
\pagebreak
\appendix

\section{Detailed Performance Across CLP Spectrum}
\label{appx:full_clp_fig}
\begin{figure}[h]
    \centering
    \includegraphics[width=1\textwidth]{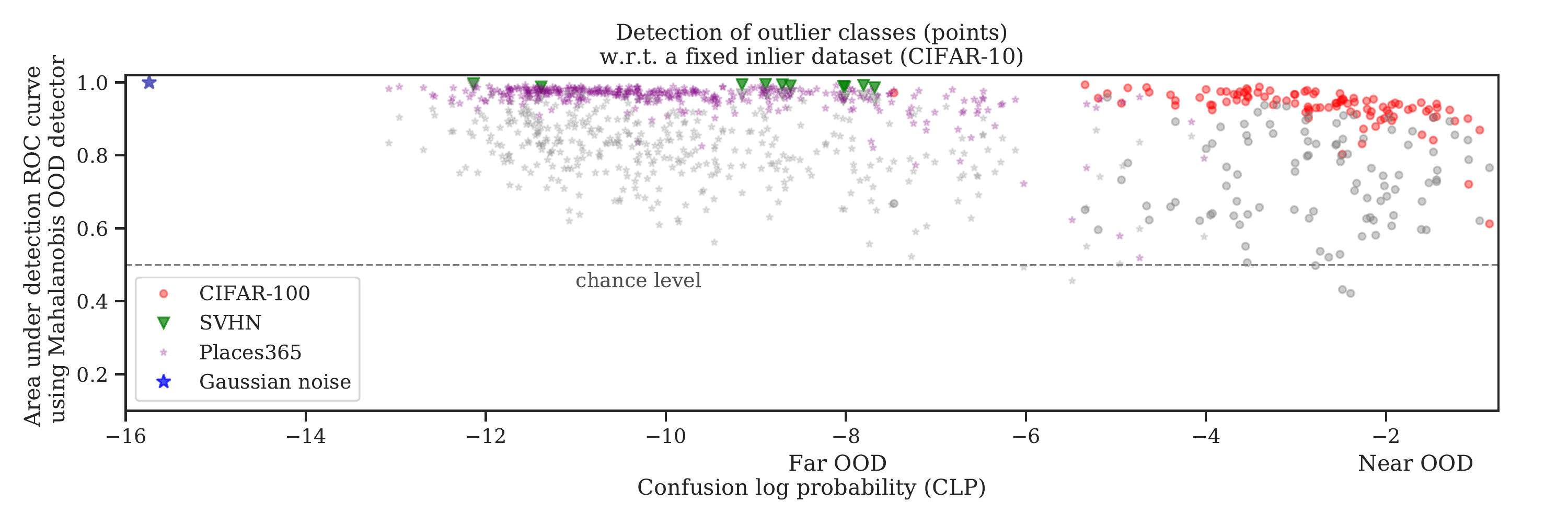}
    \caption{
    Each point represents performance at detecting one of the classes in CIFAR-100, SVHN, Places365 and Gaussian noise as outliers with respect to a network trained on CIFAR-10. Colored and gray markers correspond to performance from our best model with contrastive training, and our best model without contrastive training respectively. The classes are sorted by increasing similarity to the inlier classes as given by the class-wise confusion log probability (CLP). Contrastive training improves OOD detection results across the board, particularly in the near OOD regime, where the outlier and inlier classes are highly similar.}
    \label{fig:fullspectrum_appx}
\end{figure}

In Fig.~\ref{fig:fullspectrum_appx}, we show the class-wise OOD detection performance for the full CLP spectrum with and without contrastive training with $\Data_{\text{in}}$ being CIFAR-10 and $\Data_{\text{out}}$ being the combination of CIFAR-100, SVHN, Places365 and Gaussian noise test samples. 
Details about the training process for CLP computation are given in Sec.~\ref{sec:clp_training}.

\section{Implementation Details}
\label{appx:implementation}
We use a Resnet-50 with a $3\times$ width multiplier for all experiments. We first pretrain the model with a batch size of 2048 for 1000 epochs using only the contrastive loss and then finetune with a joint supervised and contrastive loss for 100 epochs in case of CIFAR-10 and 200 epochs for CIFAR-100. We use a supervised loss multiplier of $\lambda=100$ during the finetuning stage. The models are trained using the LARS \cite{you2017large} optimizer with momentum 0.9 and weight decay $1 \times 10^{-6}$. Furthermore, we use an initial learning rate of 1.0, with a linear warmup for the first 30 epochs followed by a cosine decay schedule without restarts following \citep{loshchilov2016sgdr} for both stages. For label smoothing, we use $\alpha=0.01$ for CIFAR-10 as $\mathcal{D}_{\text{in}}$ and $\alpha=0.1$ for CIFAR-100 as $\mathcal{D}_{\text{in}}$. 

The data augmentation operation $T$ as described in Section \ref{sec:replearn}  follows \citep{simclr}, which is a sequence of random cropping followed by a random left-right flip and random color distortion.

\section{CLP Training Details}
\label{appx:clp_train}
\label{sec:clp_training}For computing CLP scores, we trained an ensemble of five ResNet-34 model instances. In order to accurately estimate class confusions, we would ideally like to train classifiers on a dataset that is representative of all possible images. As an approximation, we independently train each of the model instances on the union of five datasets: CIFAR-10, SVHN, CIFAR-100, Places365 and independent Gaussian noise. The entire collection has $486$ classes. We used a training batch size of $1024$, and ensured that examples from each of the $486$ classes are uniformly sampled in a batch. We used SGD with momentum of 0.9 for training the models. The models were trained for 500 epochs with a cosine decay learning rate schedule initialized at 0.2. A weight decay parameter of $10^{-6}$ was used for regularization. Each of the model instances only differ in the random initializations of the weights.

Once all the model instances are trained, we use them to compute CLP between any given dataset pair. As a specific example, let us consider the case where the inlier dataset $\Data_{\text{in}}$ is CIFAR-10 and the outlier dataset $\Data_{\text{out}}$ is Places365, with 10 and 365 classes respectively. To calculate the CLP score 
of Places365 (with respect to CIFAR-10), we compute, for each of the $5$ model instances, the softmax output for all the test examples in Places365. The outputs of the model ensemble instances are averaged to have a $1 \times 486$ vector output, where $486$ is the total number of classes (in the union of all datasets). We compute the total probability of the $10$ outputs corresponding to CIFAR-10 classes as $\Data_{\text{in}}$ and report the log of this probability as CLP as an estimate of confusing a Places365 example with a CIFAR-10 example.

\begin{wrapfigure}[16]{r}{0.5\textwidth}
    \vspace{-0.5cm}
    \centering
    \includegraphics[width=0.45\textwidth]{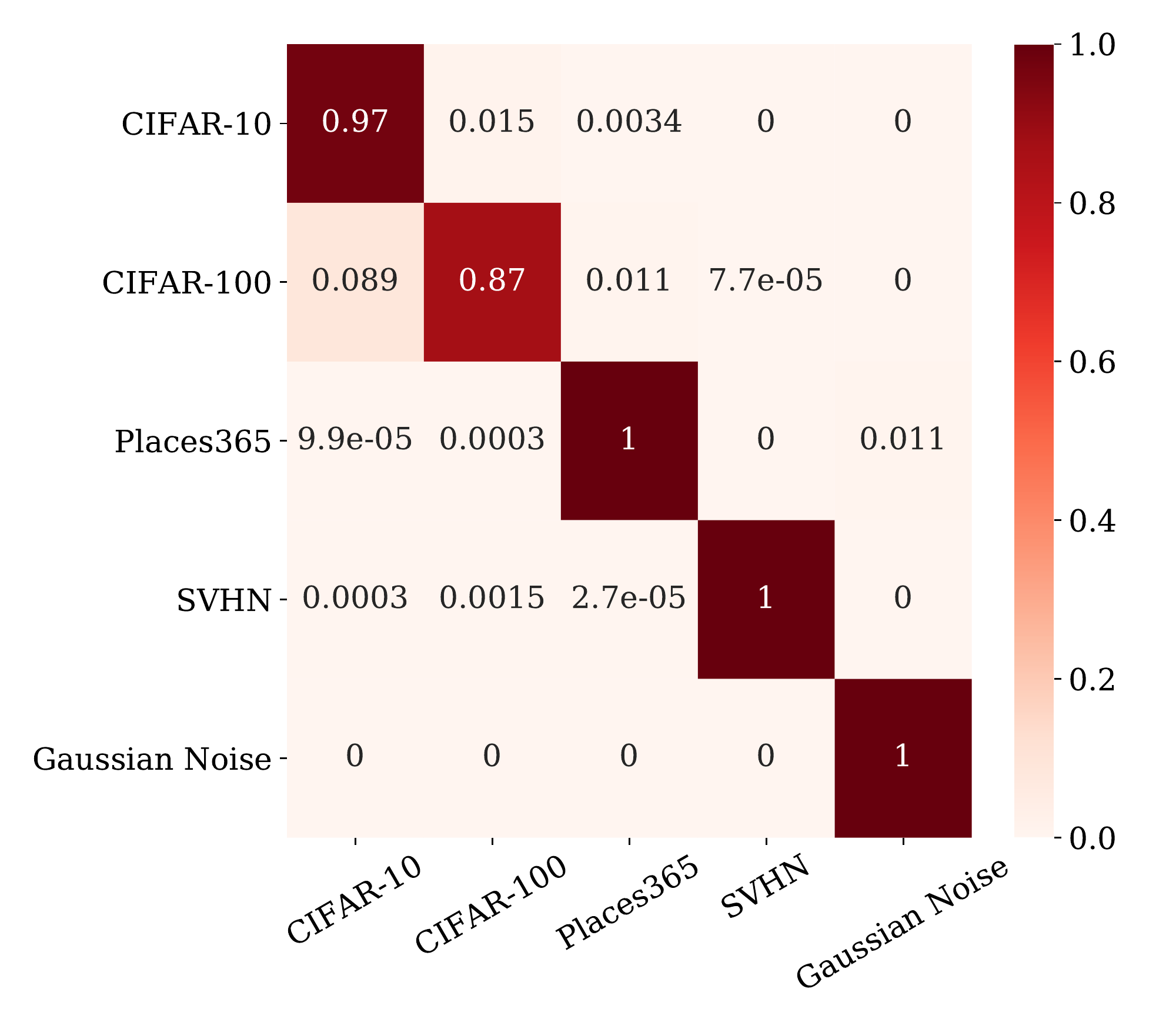}
    \vspace{-0.2cm}
    \caption{
    Confusion matrix of the classifier ensemble using the combination of the five reported datasets. Rows and columns represent the true labels and the predictions respectively.}
    \label{fig:conf_mat}
\end{wrapfigure}

The confusion matrix of the our model is shown in Fig.~\ref{fig:conf_mat}, by combining all the classes of a given dataset. The rows indicate the true labels and columns indicate the predictions. We observe that CIFAR-10 and CIFAR-100 confuses more among themselves compared to datasets: Places365, SVHN and Gaussian noise, which are far OOD to them. Also note that these datasets have almost $100\%$ classification accuracy. Thus the relative ordering of these far OOD datasets from in-distribution CIFAR-10/100 is only as good as the state-of-the art in calibrated uncertainty.

\section{Qualitative Analysis of CLP}
\label{appx:dendogram}
\subsection*{Visualization of Class Similarities}
\label{sec:dendrogram}
To qualitatively ascertain our CLP estimates, we use the model jointly trained on CIFAR-10, CIFAR-100, SVHN, Places365 and Gaussian noise resulting in a 486-way classifier. We compute a confusion probability for each pair of classes $i,j$ as
\begin{align}\label{eq:clp}
        u_{i\rightarrow j} &= 
    \frac{1}{|\mathcal{D}_{\text{test},j}|}
    \sum_{\obs \in \mathcal{D}_{\text{test},j}} 
    \!\!\!\!\!c_i(\obs)\;,
\end{align}
using the expected probability $c_i(\obs)$ of a test sample to be predicted as class $i$ (Eq. \eqref{eqn:simm_score} in the main paper), and the test set for class $j$,  $\mathcal{D}_{\text{test},j}$.
We translate the probabilities into symmetric distances using
\begin{equation}
    d(i,j) = \sqrt{-\log\left(\frac{1}{2} \left(u_{i\rightarrow j} + u_{j\rightarrow i}\right)\right)}\;.
\end{equation}

\begin{figure}[hp!]
    \centering
    \includegraphics[height=0.95\textheight]{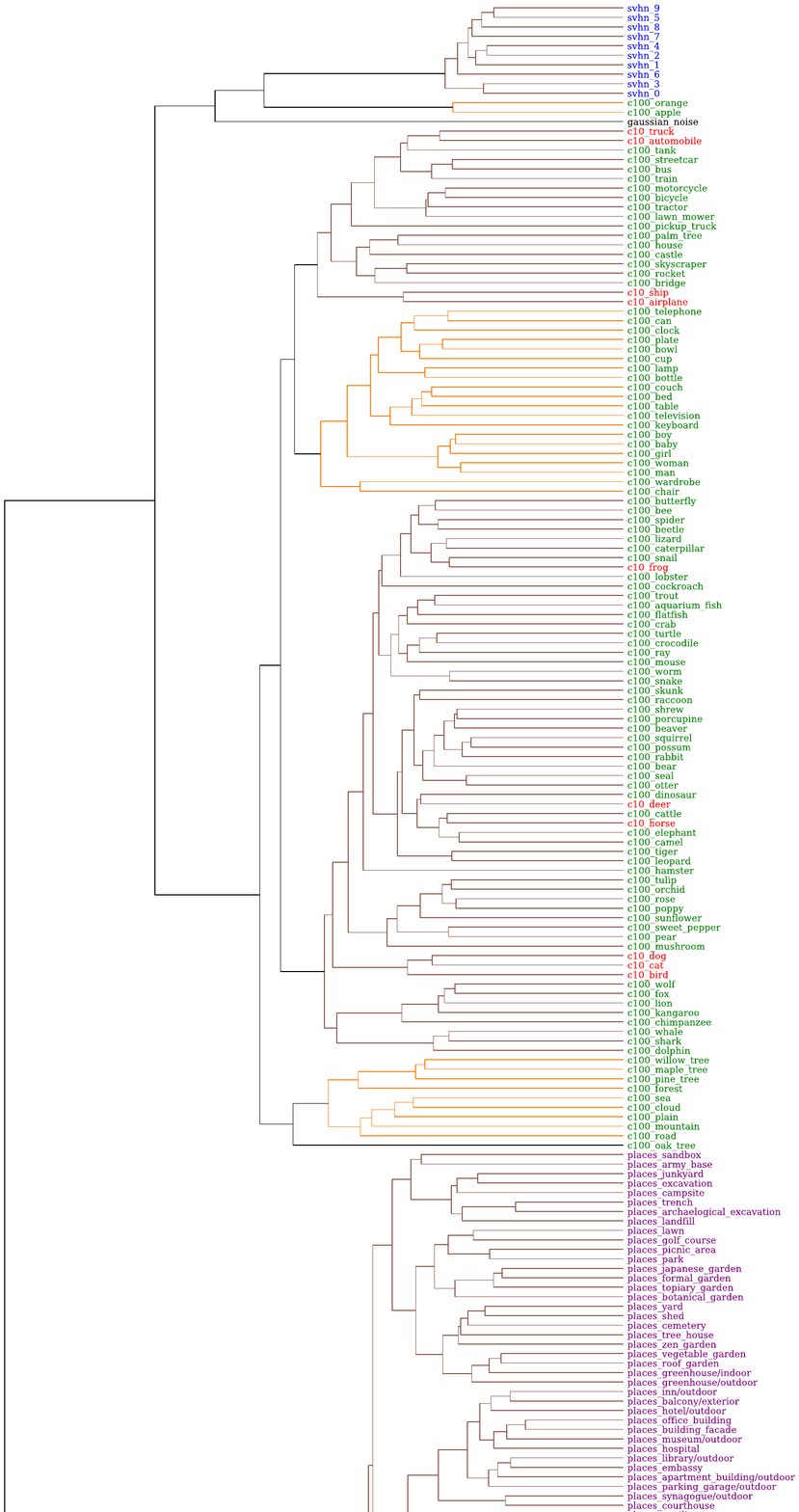}
    \caption{\textbf{Qualitative Analysis of CLP.} Three-part dendrogram plot of the  classes of CIFAR-10 (red), CIFAR-100 (green), SVHN (blue), Places365 (purple) and Gaussian noise (black) based on the expected confusion matrix combining the datasets.}
    \label{fig:dendrogram}
\end{figure}

We then use these pairwise distances to perform hierarchical agglomerative clustering (with `average' linkage).
\autoref{fig:dendrogram} shows the dendrogram. We observe that the relationships captured by this distance measure are a good representation of the visual similarity of the classes: all the SVHN classes (shown in blue) are well separated from the CIFAR-10 and CIFAR-100 classes indicating that it is a far OOD dataset. CIFAR-10 (shown in red) and CIFAR-100 classes (shown in green) on the other hand are frequently confused, indicating that the two datasets are closer to one another than to SVHN. The Places365 dataset builds a separate cluster (shown in purple). Surprisingly the CIFAR-100 classses `orange' and `apple' are quite dissimilar to the remaining CIFAR-100 and CIFAR-10 classes and get clustered closer to the house numbers. 

The individual classes within the data sets that get clustered together match very well our impression of ``visual similarity''. We take this as another confirmation that our proposed CLP metric can be used to quantify the difficulty of OOD tasks.

\begin{figure}[hp!]
    \centering
    \hspace{-2.2cm} \includegraphics[height=\textheight]{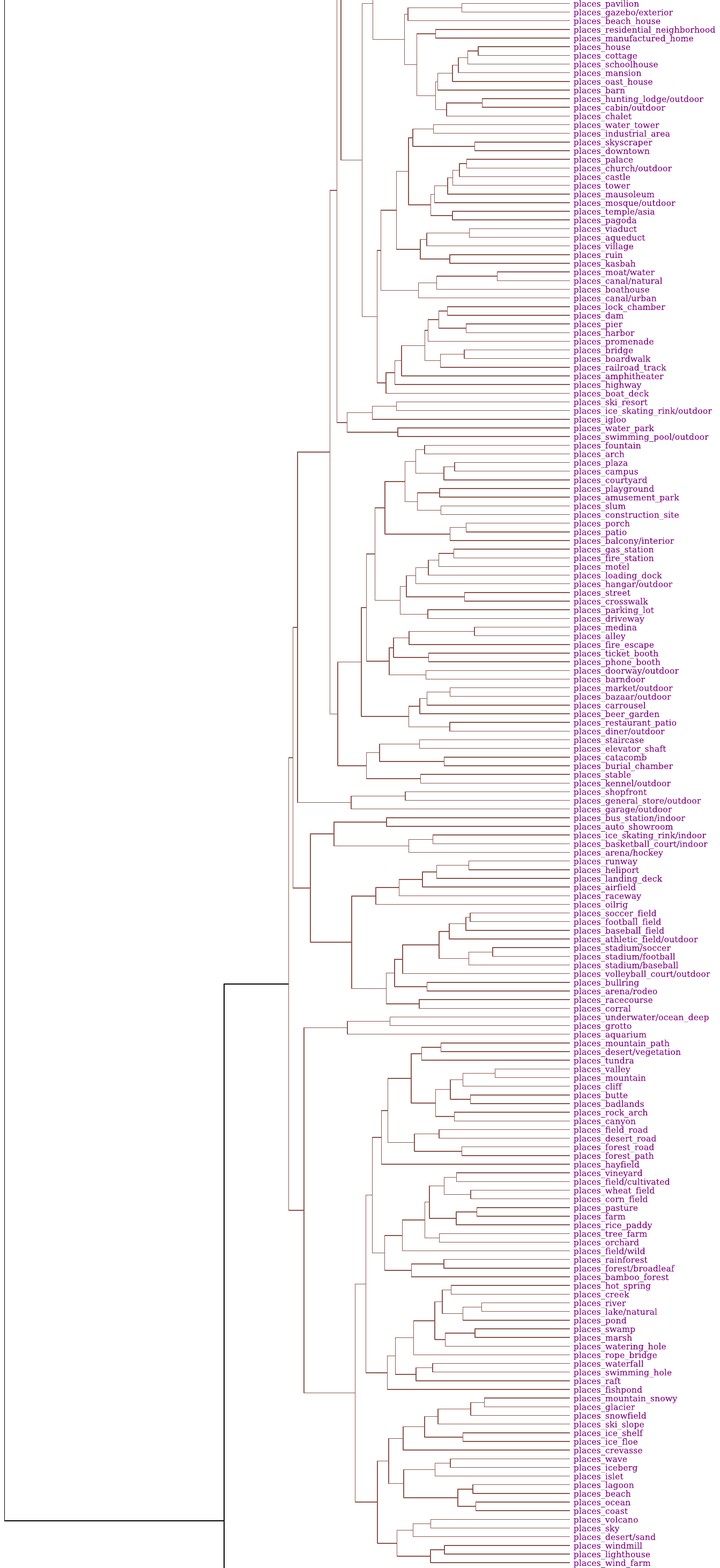}
\end{figure}

\begin{figure}[hp!]
    \includegraphics[height=\textheight]{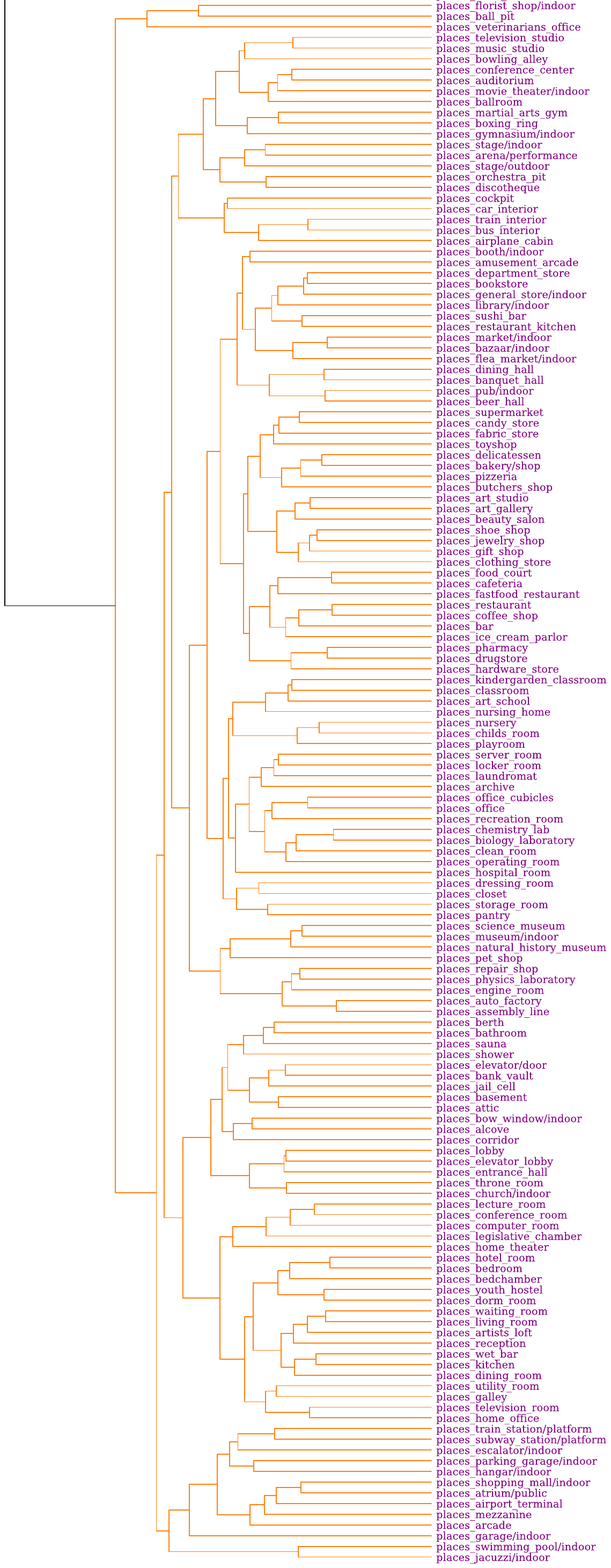}
\end{figure}
\pagebreak

\section{Additional Results}
\label{appx:additional_res}
As an extension to Table 1, we report additional metrics for models trained with our methods. We chose to report the whole
\begin{itemize}
    \item \textbf{Area Under Receiver Operating Characteristic (AUROC)} corresponds to the area under the Receiver Operating Characteristic curve, which plots the True Positive Rate (TPR) as a function of the False Positive Rate (FPR). This metric has the advantage of not requiring to choose the choice of a threshold and evaluating the OOD detector in all range of operations. It can be interpreted as the probability that, when picking an out of distribution sample and an in-distribution sample, the in-distribution sample gets considered as more in-distribution that the OOD sample.
    \item \textbf{Area Under Precision Recall (AUPR)} is a closely related metric. It corresponds to the area under the Precision-Recall Curve, which plots Precision against Recall.
    \item \textbf{False Positive Rate at 95\% True Positive Rate (FPR@95\%TPR)} is the probability that an OOD example is correctly identified when the true positive rate (TPR) is 95\%. TPR is computed as $TPR=TP/(TP+FN)$, where TP and FN denote true positive and false negative respectively. 
    \item \textbf{Detection Accuracy (DtAcc)} is a measurement of the maximum classification accuracy that we can achieve between in-distribution and out-of-distribution examples, by choosing the optimal threshold.
\end{itemize}

We also report results on additional out of distribution datasets.
\begin{itemize}
 \item \textbf{Gaussian noise} is a synthetic dataset composed of $32\times32$ pixel random images where the intensity of each pixel is sampled from a gaussian distribution with mean 0.5 and scale 0.25.
 \item \textbf{Places365}~\citep{zhou2017places} is a dataset containing scene photographs. We rescale them from their original size down to $32\times32$ to match the size of our in-distribution images.
\end{itemize}

\begin{table}[h]
\centering
\caption{Additional out-of-distribution detection results for our proposed method. All results are an average over five independent runs. We report results both for our method and the baseline that does not include label smoothing or contrastive training.}
\begin{tabular}{@{}clccccc@{}}

\toprule
& & & \multicolumn{4}{c}{Baseline / Ours}\\
\cmidrule{4-7}
$\mathcal{D}_{\text{in}}$ & $\mathcal{D}_{\text{out}}$           & CLP range & FPR@95\%TPR $\downarrow$ & AUROC $\uparrow$& AUPR $\uparrow$& DtAcc $\uparrow$\\ \midrule
\parbox[t]{2mm}{\multirow{4}{*}{\rotatebox[origin=c]{90}{\small CIFAR-10}}} 
 & CIFAR-100      & [-7.4 to -0.8] & 67.1 / 39.9 & 81.3 / 92.9  & 81.2 / 93.7 & 73.8 / 85.9\\
 & SVHN           & [-12.3 to -7.6] & 20.5 / 2.8  & 96.2 / 99.5  & 96.6 / 99.6 & 89.5 / 96.7\\
 & Places365      & [-13.1 to -4.0] & 69.2 / 24.3 & 82.0 / 95.3  & 82.6 / 95.3 & 74.7 / 89.1\\
 & Gaussian noise & -15.7 & 3.4 / 0.3 & 99.1 / 100  & 99.4 / 100 & 97.1 / 99.2\\
\midrule
 \parbox[t]{2mm}{\multirow{4}{*}{\rotatebox[origin=c]{90}{\small CIFAR-100}}}
 & CIFAR-10       & [-4.5 to -2.6] & 93.1 / 81.8 & 63.9 / 78.3 & 68.1 / 80.2 & 60.8 / 72.0\\
 & SVHN           & [-6.4 to -8.6] & 56.5 / 23.3 & 87.8 / 95.6 & 88.9 / 95.8 & 80.1 / 89.2\\
 & Places365      & [-12.1 to -4.4] & 96.4 / 69.2 & 51.8 / 82.0 & 58.3 / 82.6 & 55.1 / 78.4\\
 & Gaussian noise & -15.1 & 97.3 / 0.2  & 53.9 / 99.9 & 65.9 / 99.9 & 63.4 / 98.0\\
 \bottomrule
\end{tabular}
\label{tab:additional_res}
\end{table}

\section{Additional Ablations}
\label{appx:additional_ablation}
\textbf{Sensitivity to supervised loss multiplier.} We investigate the sensitivity of our models to the supervised loss multiplier parameter $\lambda$ in our objective function defined in Section \ref{sec:replearn} which controls the strength of the supervised loss during the finetuning stage.  We consider $\lambda\in[0, 1, 10, 100, 1000]$ for our experiments. We observe that increasing the loss multiplier leads to particularly significant improvement on the near OOD dataset pair with $\Data_{\text{in}}$ as CIFAR-100 and $\Data_{\text{out}}$ as CIFAR-10 but gains are marginal beyond $\lambda=100$. We conjecture that the large loss ratios are necessary due to the difference in scale of the supervised and contrastive loss.

\begin{table}[h]
  \caption{\textbf{Ablation study of weight between supervised loss and contrastive loss}}
\label{tab:trainingmodes}
\small
\centering
\begin{tabular}{l@{\hskip .2in}cccccc}
\toprule
   $\lambda$ & $\Data_{\text{in}}$ = CIFAR-100 & CIFAR-100 & CIFAR-100      & CIFAR-10  & CIFAR-10 & CIFAR-10 \\
   & $\Data_{\text{out}}$ = CIFAR-10 & SVHN      & Places365 & CIFAR-100 & SVHN     & Places365\\
  \midrule
   0.1  & $72.1 \pm 0.6$ & $94.5 \pm 3.6$ & $86.8 \pm 1.0$  & $88.3 \pm 0.1$ & $99.3 \pm 0.08$ & $91.4 \pm 0.4$ \\
   1    & $77.1 \pm 0.3$ & $95.9 \pm 0.3$ & $87.3 \pm 0.3$ & $91.9 \pm 0.1$ & $99.4 \pm 0.1$ & $94.1 \pm 0.3$\\
   10   & $77.9 \pm 0.3$ & $95.5 \pm 0.6$ & $86.1 \pm 0.5$ & $92.7 \pm 0.1$ & $99.4 \pm 0.04$ & $94.4 \pm 0.2$\\
   100  & $78.6 \pm 0.5$ & $95.3 \pm 0.4$ & $86.2 \pm 0.1$ & $92.9 \pm 0.2$ & $99.5 \pm 0.1$  & $94.5 \pm 0.5$\\
   1000 & $77.9 \pm 0.1$ & $95.9 \pm 0.3$ & $85.9 \pm 0.4$ & $92.9 \pm 0.2$ & $99.3 \pm 0.1$  & $95.0 \pm 0.4$\\
  \bottomrule
\end{tabular}
\end{table}

\textbf{Impact of increasing model capacity.}
Our reference model is a wide ResNet-50 with a multiplier of 3. 
We investigate the impact of model capacity by training models for other width multipliers, ranging from 1 (the default ResNet-50) to 4.
We observe that decreasing the ResNet-50 width to 1 leads to a significant drop on both near and far OOD dataset pairs. On the other hand, increasing the width to 4 does not result in improvement on OOD detection performance. We hypothesize that using contrastive training mandates the need for a higher capacity model in order to capture a richer representation with general task-agnostic features, and thus to achieve optimal performance compared to a supervised only baseline.

\begin{table}[h]
  \caption{\textbf{Ablation study of model capacity}}
\label{tab:trainingmodes}
\small
\centering
\begin{tabular}{l@{\hskip .2in}cccccc}
\toprule
   Width & $\Data_{\text{in}}$ = CIFAR-100 & CIFAR-100 & CIFAR-100 & CIFAR-10  & CIFAR-10 & CIFAR-10 \\
   & $\Data_{\text{out}}$ = CIFAR-10 & SVHN & Places365 & CIFAR-100 & SVHN & Places365\\
  \midrule
   1 & $74.2 \pm 0.4$ & $96.5 \pm 0.4$ & $85.4 \pm 0.7$& $89.8 \pm 0.2$ & $97.8 \pm 0.3$  & $93.9 \pm 0.7$  \\
   2 & $77.1 \pm 0.6$ & $96.4 \pm 0.3$ & $86.8 \pm 0.2$& $92.7 \pm 0.1$ & $99.2 \pm 0.2$  & $94.6 \pm 0.3$ \\
   3 & $78.3 \pm 0.3$ & $95.4 \pm 0.1$ & $85.3 \pm 0.5$& $92.9 \pm 0.2$ & $99.5 \pm 0.1$  & $94.7 \pm 0.3$\\
   4 & $78.9 \pm 0.2$ & $95.4 \pm 0.2$ & $85.9 \pm 0.9$& $92.6 \pm 0.2$ & $99.3 \pm 0.04$ & $94.3 \pm 0.4$ \\
  \bottomrule
\end{tabular}
\end{table}

\textbf{Effect of temperature parameter of the contrastive loss.}
Finally, we run experiments to understand the importance of the temperature parameter $\tau$ used in the contrastive loss. We consider $\tau$ values $\in[0.01, 0.1, 0.5, 1, 2]$ for our experiments. Unlike \citep{simclr} which obtains optimal performance with a $\tau$ of 0.1, we find that higher temperature leads to improved performance for the OOD detection task across the board. The optimal performance is obtained with a $\tau$ of 1.
\begin{table}[h]
  \caption{\textbf{Ablation study of contrastive loss temperature}}
\label{tab:trainingmodes}
\small
\centering
\begin{tabular}{l@{\hskip .2in}cccccc}
\toprule
   $\tau$ & $\Data_{\text{in}}$ = CIFAR-100 & CIFAR-100 & CIFAR-100      & CIFAR-10  & CIFAR-10 & CIFAR-10 \\
   & $\Data_{\text{out}}$ = CIFAR-10 & SVHN      & Places365 & CIFAR-100 & SVHN     & Places365\\
  \midrule
   0.01 & $57.8 \pm 3.0$ & $52.2 \pm 3.9$ & $62.5 \pm 1.9$ & $49.5 \pm 4.0$  & $48.1 \pm 14.5$ & $54.1 \pm 2.2$ \\
   0.1  & $77.4 \pm 0.3$ & $94.4 \pm 0.7$ & $84.5 \pm 1.3$ & $91.8 \pm 0.2$  & $99.3 \pm 0.03$ & $94.8 \pm 0.4$\\
   0.5  & $77.1 \pm 0.1$ & $96.6 \pm 0.1$ & $86.7 \pm 0.2$ & $92.9 \pm 0.06$ & $99.5 \pm 0.1$  & $94.9 \pm 0.1$ \\
   1    & $78.7 \pm 0.5$ & $95.7 \pm 0.5$ & $86.6 \pm 0.5$ & $92.9 \pm 0.1$  & $99.5 \pm 0.04$ & $94.7 \pm 0.1$\\
   2    & $78.6 \pm 0.2$ & $96.4 \pm 0.5$ & $84.6 \pm 0.3$ & $91.9 \pm 0.2$  & $99.0 \pm 0.1$  & $93.4 \pm 0.4$\\
  \bottomrule
\end{tabular}
\end{table}

\end{document}

%% file: defs.tex
\newcommand{\cls}{y}

\newcommand{\obs}{\mathbf{x}}

\newcommand{\lat}{\mathbf{z}}

\newcommand{\Data}{\mathcal{D}}

\newcommand{\phat}{\hat{p}}

\newcommand{\projclass}{g_\phi}
\newcommand{\projcontr}{h_\nu}

\newcommand{\trans}{T}
\DeclareMathOperator{\cossim}{{sim}}

%% file: arxiv_submission.bbl
\begin{thebibliography}{38}
\providecommand{\natexlab}[1]{#1}
\providecommand{\url}[1]{\texttt{#1}}
\expandafter\ifx\csname urlstyle\endcsname\relax
  \providecommand{\doi}[1]{doi: #1}\else
  \providecommand{\doi}{doi: \begingroup \urlstyle{rm}\Url}\fi

\bibitem[Bachman et~al.(2019)Bachman, Hjelm, and Buchwalter]{amdim}
Bachman, P., Hjelm, R.~D., and Buchwalter, W.
\newblock Learning representations by maximizing mutual information across
  views.
\newblock In \emph{Advances in Neural Information Processing Systems}, 2019.

\bibitem[Blundell et~al.(2015)Blundell, Cornebise, Kavukcuoglu, and
  Wierstra]{blundell2015weight}
Blundell, C., Cornebise, J., Kavukcuoglu, K., and Wierstra, D.
\newblock Weight uncertainty in neural networks.
\newblock \emph{arXiv preprint arXiv:1505.05424}, 2015.

\bibitem[Chen et~al.(2020)Chen, Kornblith, Norouzi, and Hinton]{simclr}
Chen, T., Kornblith, S., Norouzi, M., and Hinton, G.
\newblock A simple framework for contrastive learning of visual
  representations.
\newblock In \emph{International Conference on Machine Learning}, 2020.

\bibitem[Chen et~al.(2018)Chen, Shen, Jin, and Wang]{chen2018variational}
Chen, W., Shen, Y., Jin, H., and Wang, W.
\newblock A variational dirichlet framework for out-of-distribution detection.
\newblock \emph{arXiv preprint arXiv:1811.07308}, 2018.

\bibitem[Choi et~al.(2018)Choi, Jang, and Alemi]{choi2018waic}
Choi, H., Jang, E., and Alemi, A.~A.
\newblock Waic, but why? generative ensembles for robust anomaly detection.
\newblock \emph{arXiv preprint arXiv:1810.01392}, 2018.

\bibitem[DeVries \& Taylor(2018)DeVries and Taylor]{devries2018learning}
DeVries, T. and Taylor, G.~W.
\newblock Learning confidence for out-of-distribution detection in neural
  networks.
\newblock \emph{arXiv preprint arXiv:1802.04865}, 2018.

\bibitem[Gal \& Ghahramani(2016)Gal and Ghahramani]{gal2016dropout}
Gal, Y. and Ghahramani, Z.
\newblock Dropout as a bayesian approximation: Representing model uncertainty
  in deep learning.
\newblock In \emph{International Conference on Machine Learning}, pp.\
  1050--1059, 2016.

\bibitem[He et~al.(2016)He, Zhang, Ren, and Sun]{he2016deep}
He, K., Zhang, X., Ren, S., and Sun, J.
\newblock Deep residual learning for image recognition.
\newblock In \emph{Proceedings of the IEEE conference on Computer Vision and
  Pattern Recognition}, pp.\  770--778, 2016.

\bibitem[He et~al.(2019)He, Fan, Wu, Xie, and Girshick]{moco}
He, K., Fan, H., Wu, Y., Xie, S., and Girshick, R.
\newblock Momentum contrast for unsupervised visual representation learning.
\newblock \emph{arXiv preprint arXiv:1911.05722}, 2019.

\bibitem[H{\'{e}}naff et~al.(2020)H{\'{e}}naff, Srinivas, Fauw, Razavi,
  Doersch, Eslami, and van~den Oord]{cpc}
H{\'{e}}naff, O.~J., Srinivas, A., Fauw, J.~D., Razavi, A., Doersch, C.,
  Eslami, S. M.~A., and van~den Oord, A.
\newblock Data-efficient image recognition with contrastive predictive coding.
\newblock In \emph{International Conference on Machine Learning}, 2020.

\bibitem[Hendrycks \& Gimpel(2017)Hendrycks and Gimpel]{hendrycks2016baseline}
Hendrycks, D. and Gimpel, K.
\newblock A baseline for detecting misclassified and out-of-distribution
  examples in neural networks.
\newblock In \emph{International Conference on Learning Representations}, 2017.

\bibitem[Hendrycks et~al.(2019{\natexlab{a}})Hendrycks, Mazeika, and
  Dietterich]{hendrycks2018deep}
Hendrycks, D., Mazeika, M., and Dietterich, T.
\newblock Deep anomaly detection with outlier exposure.
\newblock In \emph{International Conference on Learning Representations},
  2019{\natexlab{a}}.

\bibitem[Hendrycks et~al.(2019{\natexlab{b}})Hendrycks, Mazeika, Kadavath, and
  Song]{hendrycks2019using}
Hendrycks, D., Mazeika, M., Kadavath, S., and Song, D.
\newblock Using self-supervised learning can improve model robustness and
  uncertainty.
\newblock In \emph{Advances in Neural Information Processing Systems}, pp.\
  15637--15648, 2019{\natexlab{b}}.

\bibitem[Kamoi \& Kobayashi(2020)Kamoi and Kobayashi]{kamoi2020mahalanobis}
Kamoi, R. and Kobayashi, K.
\newblock Why is the mahalanobis distance effective for anomaly detection?
\newblock \emph{arXiv preprint arXiv:2003.00402}, 2020.

\bibitem[Krizhevsky(2009)]{Krizhevsky09learningmultiple}
Krizhevsky, A.
\newblock Learning multiple layers of features from tiny images.
\newblock Technical report, 2009.

\bibitem[Lakshminarayanan et~al.(2017)Lakshminarayanan, Pritzel, and
  Blundell]{lakshminarayanan2017simple}
Lakshminarayanan, B., Pritzel, A., and Blundell, C.
\newblock Simple and scalable predictive uncertainty estimation using deep
  ensembles.
\newblock In \emph{Advances in neural information processing systems}, pp.\
  6402--6413, 2017.

\bibitem[Lee et~al.(2018{\natexlab{a}})Lee, Lee, Lee, and
  Shin]{lee2017training}
Lee, K., Lee, H., Lee, K., and Shin, J.
\newblock Training confidence-calibrated classifiers for detecting
  out-of-distribution samples.
\newblock In \emph{International Conference on Learning Representations},
  2018{\natexlab{a}}.

\bibitem[Lee et~al.(2018{\natexlab{b}})Lee, Lee, Lee, and Shin]{lee2018simple}
Lee, K., Lee, K., Lee, H., and Shin, J.
\newblock A simple unified framework for detecting out-of-distribution samples
  and adversarial attacks.
\newblock In \emph{Advances in Neural Information Processing Systems}, pp.\
  7167--7177, 2018{\natexlab{b}}.

\bibitem[Liang et~al.(2018)Liang, Li, and Srikant]{liang2018enhancing}
Liang, S., Li, Y., and Srikant, R.
\newblock Enhancing the reliability of out-of-distribution image detection in
  neural networks.
\newblock In \emph{International Conference on Learning Representations}, 2018.

\bibitem[Loshchilov \& Hutter(2017)Loshchilov and Hutter]{loshchilov2016sgdr}
Loshchilov, I. and Hutter, F.
\newblock Sgdr: Stochastic gradient descent with warm restarts.
\newblock In \emph{International Conference on Learning Representations}, 2017.

\bibitem[Malinin \& Gales(2018)Malinin and Gales]{malinin2018predictive}
Malinin, A. and Gales, M.
\newblock Predictive uncertainty estimation via prior networks.
\newblock In \emph{Advances in Neural Information Processing Systems}, pp.\
  7047--7058, 2018.

\bibitem[Masana et~al.(2018)Masana, Ruiz, Serrat, van~de Weijer, and
  Lopez]{masana2018metric}
Masana, M., Ruiz, I., Serrat, J., van~de Weijer, J., and Lopez, A.~M.
\newblock Metric learning for novelty and anomaly detection.
\newblock In \emph{British Machine Vision Conference}, 2018.

\bibitem[Mohseni et~al.(2020)Mohseni, Pitale, Yadawa, and
  Wang]{mohseni2020self}
Mohseni, S., Pitale, M., Yadawa, J., and Wang, Z.
\newblock Self-supervised learning for generalizable out-of-distribution
  detection.
\newblock In \emph{Association for the Advancement of Artificial Intelligence},
  2020.

\bibitem[M{\"u}ller et~al.(2019)M{\"u}ller, Kornblith, and
  Hinton]{muller2019does}
M{\"u}ller, R., Kornblith, S., and Hinton, G.~E.
\newblock When does label smoothing help?
\newblock In \emph{Advances in Neural Information Processing Systems}, pp.\
  4696--4705, 2019.

\bibitem[Nalisnick et~al.(2019{\natexlab{a}})Nalisnick, Matsukawa, Teh, Gorur,
  and Lakshminarayanan]{nalisnick2018deep}
Nalisnick, E., Matsukawa, A., Teh, Y.~W., Gorur, D., and Lakshminarayanan, B.
\newblock Do deep generative models know what they don't know?
\newblock In \emph{International Conference on Learning Representations},
  2019{\natexlab{a}}.

\bibitem[Nalisnick et~al.(2019{\natexlab{b}})Nalisnick, Matsukawa, Teh, Gorur,
  and Lakshminarayanan]{nalisnick2019hybrid}
Nalisnick, E., Matsukawa, A., Teh, Y.~W., Gorur, D., and Lakshminarayanan, B.
\newblock Hybrid models with deep and invertible features.
\newblock In \emph{International Conference on Machine Learning},
  2019{\natexlab{b}}.

\bibitem[Netzer et~al.(2011)Netzer, Wang, Coates, Bissacco, Wu, and
  Ng]{Netzer2011}
Netzer, Y., Wang, T., Coates, A., Bissacco, A., Wu, B., and Ng, A.~Y.
\newblock Reading digits in natural images with unsupervised feature learning.
\newblock In \emph{Advances in Neural Information Processing Systems}, 2011.

\bibitem[Nguyen et~al.(2015)Nguyen, Yosinski, and Clune]{nguyen2015deep}
Nguyen, A., Yosinski, J., and Clune, J.
\newblock Deep neural networks are easily fooled: High confidence predictions
  for unrecognizable images.
\newblock In \emph{Proceedings of the IEEE conference on Computer Vision and
  Pattern Recognition}, pp.\  427--436, 2015.

\bibitem[Recht et~al.(2019)Recht, Roelofs, Schmidt, and
  Shankar]{recht2019imagenet}
Recht, B., Roelofs, R., Schmidt, L., and Shankar, V.
\newblock Do imagenet classifiers generalize to imagenet?
\newblock \emph{arXiv preprint arXiv:1902.10811}, 2019.

\bibitem[Ren et~al.(2019)Ren, Liu, Fertig, Snoek, Poplin, Depristo, Dillon, and
  Lakshminarayanan]{ren2019likelihood}
Ren, J., Liu, P.~J., Fertig, E., Snoek, J., Poplin, R., Depristo, M., Dillon,
  J., and Lakshminarayanan, B.
\newblock Likelihood ratios for out-of-distribution detection.
\newblock In \emph{Advances in Neural Information Processing Systems}, pp.\
  14680--14691, 2019.

\bibitem[Sastry \& Oore(2020)Sastry and Oore]{sastry2019detecting}
Sastry, C.~S. and Oore, S.
\newblock Detecting out-of-distribution examples with gram matrices.
\newblock In \emph{International Conference on Machine Learning}, 2020.

\bibitem[Scheirer et~al.(2012)Scheirer, de~Rezende~Rocha, Sapkota, and
  Boult]{scheirer2012toward}
Scheirer, W.~J., de~Rezende~Rocha, A., Sapkota, A., and Boult, T.~E.
\newblock Toward open set recognition.
\newblock \emph{IEEE transactions on Pattern Analysis and Machine
  Intelligence}, 35\penalty0 (7):\penalty0 1757--1772, 2012.

\bibitem[Shalev et~al.(2018)Shalev, Adi, and Keshet]{shalev2018out}
Shalev, G., Adi, Y., and Keshet, J.
\newblock Out-of-distribution detection using multiple semantic label
  representations.
\newblock In \emph{Advances in Neural Information Processing Systems}, pp.\
  7375--7385, 2018.

\bibitem[Vyas et~al.(2018)Vyas, Jammalamadaka, Zhu, Das, Kaul, and
  Willke]{vyas2018out}
Vyas, A., Jammalamadaka, N., Zhu, X., Das, D., Kaul, B., and Willke, T.~L.
\newblock Out-of-distribution detection using an ensemble of self supervised
  leave-out classifiers.
\newblock In \emph{Proceedings of the European Conference on Computer Vision},
  pp.\  550--564, 2018.

\bibitem[You et~al.(2017)You, Gitman, and Ginsburg]{you2017large}
You, Y., Gitman, I., and Ginsburg, B.
\newblock Large batch training of convolutional networks.
\newblock \emph{arXiv preprint arXiv:1708.03888}, 2017.

\bibitem[Zhang et~al.(2020)Zhang, Li, Guo, and Guo]{zhang2020hybrid}
Zhang, H., Li, A., Guo, J., and Guo, Y.
\newblock Hybrid models for open set recognition.
\newblock \emph{arXiv preprint arXiv:2003.12506}, 2020.

\bibitem[Zhou et~al.(2017)Zhou, Lapedriza, Khosla, Oliva, and
  Torralba]{zhou2017places}
Zhou, B., Lapedriza, A., Khosla, A., Oliva, A., and Torralba, A.
\newblock Places: A 10 million image database for scene recognition.
\newblock \emph{IEEE Transactions on Pattern Analysis and Machine
  Intelligence}, 2017.

\bibitem[Zisselman \& Tamar(2020)Zisselman and Tamar]{zisselman2020deep}
Zisselman, E. and Tamar, A.
\newblock Deep residual flow for novelty detection.
\newblock \emph{arXiv preprint arXiv:2001.05419}, 2020.

\end{thebibliography}
